\documentclass{article}

\usepackage{microtype}
\usepackage{graphicx}
\usepackage{subcaption}
\usepackage{booktabs} 
\usepackage{multirow}

\usepackage{hyperref}

\usepackage[main,final]{neurips2026/neurips_2026}

\usepackage{comment}

\usepackage{enumitem}

\usepackage[table]{xcolor}
\definecolor{mypink}{RGB}{255, 220, 230}  
\definecolor{mygreen}{RGB}{200, 255, 200} 

\usepackage{amsmath}
\usepackage{amssymb}
\usepackage{mathtools}
\usepackage{amsthm}
\usepackage{algorithm}
\usepackage[noend]{algpseudocode}
\usepackage{wrapfig}

\usepackage{graphicx}
\graphicspath{{./Figures/}}
\usepackage[capitalize]{cleveref}

\theoremstyle{plain}
\newtheorem{theorem}{Theorem}[section]

\newtheorem{corollary}[theorem]{Corollary}
\theoremstyle{definition}
\newtheorem{definition}[theorem]{Definition}

\theoremstyle{remark}

\usepackage[table]{xcolor}
\usepackage{colortbl} 

\definecolor{lightgreen}{RGB}{220, 255, 220}  

\newcommand{\db}[1]{\textcolor{orange}{\emph{A0: #1}}}

\title{\textsc{RCStat}: A Statistical Framework of Relative Contextualization in Transformers}

\author{%
    Debabrata Mahapatra \\
    Adobe Research\\
    Bangalore, India \\
    \texttt{dmahapatra@adobe.com}
    \And
    Shubham Agarwal \\
    UC Berkeley\\
    Berkeley, CA, USA \\
    \texttt{shubham3@berkeley.edu}
    \AND
    Apoorv Saxena \\
    Inception Labs\\
    \texttt{apoorv@inceptionlabs.ai}
    \And
    Subrata Mitra \\
    Adobe Research\\
    Bangalore, India \\
    \texttt{sumitra@adobe.com}
}

\begin{document}
\maketitle

\begin{abstract}
Identifying which tokens and activations truly influence model predictions is critical for both the efficiency and interpretability of large auto-regressive language models. Yet in practice, token importance is typically inferred from attention distributions that entangle contextual relevance with softmax-normalization effects, obscuring relative influence across tokens and heads. 
We propose $\textbf{RCStat}$, a statistical framework for quantifying contextual influence in attention mechanisms. At its core $\textbf{Relative Contextualization (RC)}$ is a random variable that measures how strongly one subset of tokens contributes to another under the model’s internal scoring scheme. RCStat admits computationally efficient bounds on expected influence that can be estimated at inference time. No retraining is required.
We apply RCStat to two tasks. For $\textbf{attribution}$, attention heads with high expected RC accurately identify the prompt spans that drive generation, enabling reliable span-level explanations. 
For $\textbf{key-value cache eviction}$, RC-based adaptive thresholding selectively evicts low-impact KV entries, substantially reducing cache size while preserving generation quality. 
Across question answering, summarization, and attribution benchmarks, RCStat achieves consistent gains, improving generation quality by 15–40\% with upto 36\% error reduction for KV eviction, and attribution accuracy by 2–16\%,. These results demonstrate that explicitly modeling contextual influence provides a principled and practical alternative to attention-based heuristics. 
\end{abstract}


\section{Introduction}
\label{sec: intro}
\begin{figure*}[t]
    \centering
    \includegraphics[width=\linewidth]{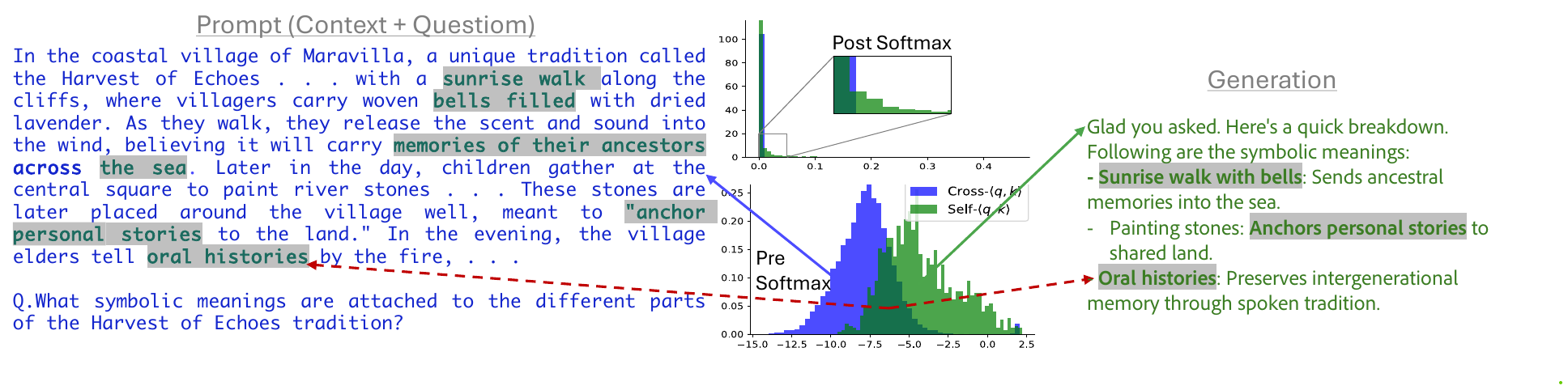} 
    \vspace{-8mm}
    \caption{
    Attribution between generated text (right) and input prompt (left) is analyzed using pre-softmax attention logits vs.\ post-softmax values. The bottom histogram highlights pre-softmax logits, separating prompt tokens (cross-$\langle q,k\rangle$) from generated tokens (self-$\langle q,k\rangle$), at the $13^\text{th}$ layer and $23^\text{th}$ head of Llama-3B-instruct.
    Tokens in the overlapping region signify common content, which is suppressed in the post-softmax histogram above. Using logit distributions, we attribute (\S~\ref{sec:attribution}): (1) prompt parts not contextualized during generation, (2) generation parts uninfluenced by the prompt, and (3) generation heavily contextualized by the prompt (red dashed arrows).
    }
    \label{fig:intro}
\end{figure*}

The transformer’s attention~\citep{10.5555/3295222.3295349} mechanism encodes contextual relationships between tokens into internal state representations. This involves raw dot-product similarity scores $\langle q, k\rangle$ of the query and key vectors, followed by softmax normalization.
Post-softmax attention weights are widely used for tasks such as attribution~\citep{yue2023automatic, phukan2024peering} and memory optimization through Key-Value (KV) cache compression~\citep{gemodel, liu2024minicache, liu2023scissorhands, li2024snapkv}.
However, this transformation introduces structural bias: it sharpens attention toward dominant tokens while flattening mid-range scores, discarding subtle contextual signals.
\cref{fig:intro} visualizes pre-softmax attention logits $(\langle q, k\rangle)$ from generated tokens to the prompt. Prompt tokens semantically relevant to the generation, i.e., carrying contextual alignment, obtain higher logit values, while unrelated prompt tokens obtain lower logits. These differences are evident in pre-softmax but are obscured in post-softmax, where normalization flattens intermediate scores and skews attention~\citep{xiao2023efficient} toward dominant or structurally favored positions (e.g., \texttt{<s>}, \texttt{</s>}), referred to as the \textit{attention sink} phenomenon~\citep{gu2024attention, xiao2023efficient}.


Such information loss becomes consequential in applications requiring fine-grained relevance estimation use-cases, leading to inaccurate token attribution \citep{li2024attributionbench}, sub-optimal KV-eviction \citep{ren2024efficacy}, etc.
While post-softmax weights represent localized attention at a specific layer, we posit that raw logits carry a dual role: they encode not only what the current layer attends to but also preserve upstream interactions, offering a richer statistical substrate for analysis.
However, the informativeness of pre-softmax logits may not be the same across all attention heads. In fact, it is observed in literature that certain heads -- often in the middle layers -- demonstrate stronger contextual activity than others \citep{phukan2024peering,skean2025layer}. Recent interpretability efforts~\citep{dunefsky2024transcoders}, such as circuit tracing \citep{ameisen2025circuit}, offer valuable but often qualitative insights into attention behavior. 
%
%
While there are some prior works~\cite{Tangetal2024, Xiaoetal2024InfLLM, Xiaoetal2025DuoAttention} that explored the usage of pre-softmax attentions, a comprehensive methodology and necessary statistical framework to extract structured insights from such unnormalized logits is missing and is needed to develop richer understanding of how such logits can influence the behavior of the LLMs. 

To addresses this gap, in this work we propose a probabilistic formalism that models relevance directly in the logit space, at different levels of granularity, enabling actionable and generalizable utilities across multiple downstream tasks.
Our framework can be used to perform head-aware attention-logit analysis by quantifying how logit distributions differ across different attention heads, and using that rank heads that shows distinct signal for that application. 

We demonstrate that this framework can seamlessly work for both attribution (finding which part of a generated action came from which part of the input context) and KV-cache compression.
Additionally, we also discover that most of heads have compressible KV-caches, while the few resistant ones provide clear and useful attribution signals.

We summarize our contributions as follows:

\vspace{-3mm}

\begin{enumerate}
[leftmargin=*, labelsep=0.5em, itemsep=0pt, parsep=0pt, topsep=0pt]
    \item We provide a statistical framework to analyze pre-softmax attention logits by formalizing the notion of contextualization as random variables that capture the relationship between two portions of a text, e.g, a part of prompt and a part of generated tokens, within an attention head. Armed with these random variables, we introduce the relative contextualization (RC) to assign relevance scores at different granularity levels: token-, chunk- and entire text.
    \item To estimate the statistics of RC, we derive an easy-to-compute, intuitive upper bound and provide an efficient algorithm to compute it, enabling quantitative use of pre-softmax attention logits in downstream tasks.
    \item In doing so, we propose \textsc{RCStat}, which
    to the best of our knowledge, is the first unified framework that brings different relevance assignment applications such as attribution of generated answer to the input context and, KV-cache compression and under a single formalism.
    
    \item We demonstrate that RC-guided attribution, selecting just 2\% of attention heads, increases span-level attribution accuracy by 2–3\% on LLaMa and 13–16\% on Qwen3, without any model retraining. We also show that RC-based KV compression improves generation quality by 15–40\% and achieves 2–5\% higher compression over prior SOTA on LLaMa, and reduces KV cache error by up to 36\% on Qwen3 for summarization and QA tasks.
    

\end{enumerate}


\section{Related Work}

\textbf{Mechanistic Interpretability of LLMs:}
Mechanistic interpretability~\citep{liu2025mechanisticinterpretabilitymeets} aims to reverse-engineer the internal workings of large language models. Circuit-tracing techniques~\citep{ameisen2025circuit, elhage2025biology}, such as those from Anthropic, have revealed neuron-level pathways and interpretable MLP circuits. Complementary efforts dissect self-attention heads~\citep{voita2019analyzing}, uncovering roles such as induction copying~\citep{mcdougall2023copy, olsson2022context} and positional tracking~\citep{dufter2022position}. Beyond this, representation-level probes in BERT and GPT leverage attribution~\citep{rahimi2025explanations} and linear classifiers~\citep{du2025gpt, rogers2021primer, chanin2023identifying} to map hidden activations to semantic features. While insightful, these methods often rely on heuristics~\citep{gu2024attention} and remain qualitative. They lack a unified quantitative framework and offer limited utility~\citep{zhao2024explainability}. 
In contrast, our method analyzes raw attention logits across heads via the statistical lens of Relative Contextualization (RC) to identify which heads are responsible for context-grounding and by how much.


\textbf{Attribution:}
Performing attribution is critical for trustworthy generation. Existing methods rely on either gradient signals or post-softmax attention weights. Gradient- and perturbation-driven methods such as Integrated Gradients~\citep{miglani2023using, sundararajan2017axiomatic}, LIME/SHAP~\citep{ribeiro2016should,lundberg2017unified}, and masking/occlusion~\citep{schinagl2022occam} trace output sensitivity back to inputs but are computationally expensive. Attention-based methods aggregate post-softmax weights across heads and layers~\citep{abnar2020quantifying} or formulate attribution metrics~\citep{chefer2021transformer}. However, averaging suppresses medium-strength token associations due to softmax normalization and the presence of sink tokens~\citep{phukan2024peering}.
The learned explainer~\citep{cohen2025learning} mitigates this by training a model to assign reliability scores to heads using labeled data. In contrast, RC computes per-head reliability scores without labels, adapting per example without any extra training.


\begin{figure*}[t]
    \centering
    \begin{subfigure}{0.25\linewidth}
        \centering
        \includegraphics[page=1,trim=28cm 16cm 28cm 16cm, clip,width=\linewidth]{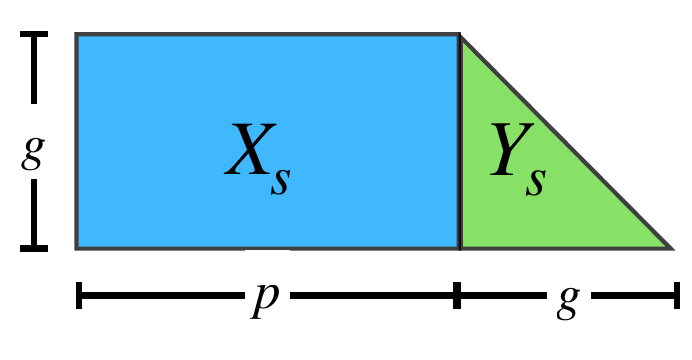}
        \caption{\ref{eq:cc} and \ref{eq:sc}}
        \label{fig:cc}
    \end{subfigure}~
    \begin{subfigure}{0.25\linewidth}
        \centering
        \includegraphics[page=2,trim=28cm 16cm 28cm 16cm, clip,width=\linewidth]{Figures/illustrations.pdf}
        \caption{Conditional CC \eqref{eq:ccc}}
        \label{fig:ccc}
    \end{subfigure}~
    \begin{subfigure}{0.25\linewidth}
        \centering
        \includegraphics[page=3,trim=28cm 16cm 28cm 16cm, clip,width=\linewidth]{Figures/illustrations.pdf}
        \caption{Conditional SC \eqref{eq:csc}}
        \label{fig:csc}
    \end{subfigure}~
    \begin{subfigure}{0.25\linewidth}
        \centering
        \includegraphics[page=4,trim=28cm 16cm 28cm 16cm, clip,width=\linewidth]{Figures/illustrations.pdf}
        \caption{Density}
        \label{fig:density}
    \end{subfigure}
    \caption{Illustration of the pre-softmax attention logits $Q^T K \in \mathbb{R}^{|s|\times |s|}$, where $s$ is sequence $s$ with prompt $p$ and generated tokens $g$, as they appear in the last $|g|$ rows the the $Q^T K$ matrix. \cref{fig:cc} shows how the logits in the last $|g|$ rows are partitioned to construct the cross-contextualization (CC) and self-contextualization (SC) random variables. \cref{fig:ccc} and \ref{fig:csc} show the logits that construct conditional CC and conditional SC. \cref{fig:density} shows their respective probability density function.
    \label{fig:illustration}}
    \vspace{-3mm}
\end{figure*}

\textbf{KV Compression:}
KV cache can occupy up to $84$\% of inference memory~\citep{hooper2024kvquant}. Reduction methods include quantization~\citep{hooper2024kvquant, liu2024kivi, lin2024matryoshkakv}, low-rank approximation~\citep{chang2024palu, dong2024get}, state merging~\citep{wang2024model, agarwal2025cache}, and eviction~\citep{li2024survey}. We focus on eviction, discarding low-value key–value pairs, which can be fixed-size or variable-size. Fixed-size methods enforce uniform budgets per head and layer; early approaches like K-norm~\citep{devoto2024simple} and StreamingLLM~\citep{xiao2023efficient} use vector norms, while attention-based methods, such as SnapKV~\citep{li2024snapkv}, TOVA~\citep{oren2024transformers}, H2O~\citep{zhang2023h2o}, rank tokens by post-softmax weights. QFilter~\citep{godey2025q} isolates signal entries via matrix decompositions, and KVPress~\citep{jegou2024kvpress} applies probabilistic heuristics. Variable-size strategies allow heterogeneous budgets: 
PyramidKV uses a fixed schedule, and Ada-KV allocates per-layer budgets by token relevance; both rely on manual budgets that risks degradation, while our method adapts budgets per head.

\textbf{Use of Pre-Softmax:}
Recent works optimize long-context inference by utilizing importance signals from pre-softmax attention logits. Quest \cite{Tangetal2024} utilizes min-max key metadata to bound and estimate these raw scores for page-level KV cache selection. Similarly, DuoAttention \cite{Xiaoetal2025DuoAttention} and InfLLM \cite{Xiaoetal2024InfLLM} reduce overhead by identifying critical retrieval components—either through head specialization or representative token lookups—that prioritize high-magnitude attention signals. However, these techniques do not provide a formalism or theoretical framework on how to distill such signals. This paper makes a significant contribution by providing this generalizable statistical framework that exploits the distributional aspects of these logits to refine the signals and assign importance to heads for a particular task.

\section{Cross, Self and Relative Contextualization}\label{sec:ccscrc}
To motivate our framework, we treat $\langle q,k\rangle$ as a random variable. This supports statistical reasoning when decisions are made over chunks rather than individual tokens, as in attribution and KV-compression.

\begin{figure*}[t]
    \centering
    \begin{subfigure}{0.24\linewidth}
        \centering
        \includegraphics[page=5,trim=28cm 16cm 28cm 16cm, clip,width=\linewidth]{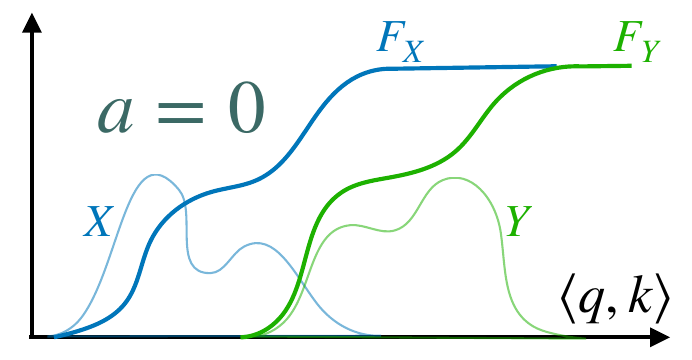}
        \caption{$a\!\leq\!\mathbb{E}[\max(X-Y, 0)]$}
        \label{fig:lb_rc_x-y}
    \end{subfigure}~
    \begin{subfigure}{0.24\linewidth}
        \centering
        \includegraphics[page=6,trim=28cm 16cm 28cm 16cm, clip,width=\linewidth]{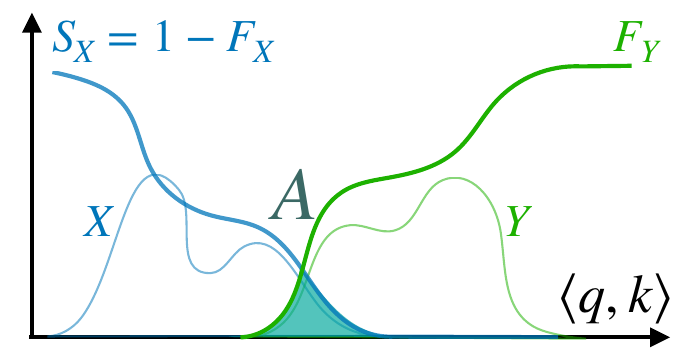}
        \caption{$\mathbb{E}[\max(X-Y, 0)]\!\leq \!A$}
        \label{fig:ub_rc_x-y}
    \end{subfigure}~
    \begin{subfigure}{0.24\linewidth}
        \centering
        \includegraphics[page=7,trim=28cm 16cm 28cm 16cm, clip,width=\linewidth]{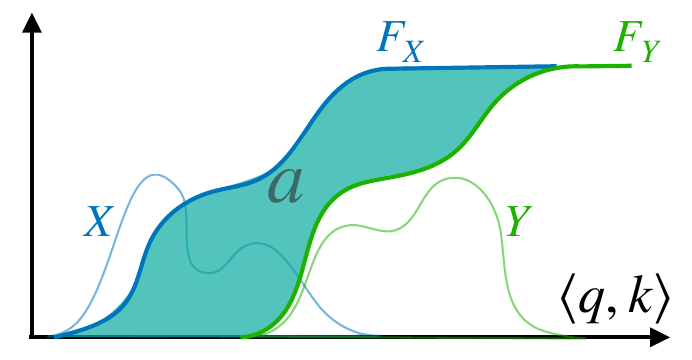}
        \caption{$a\!\leq\!\mathbb{E}[\max(Y-X, 0)]$}
        \label{fig:lb_rc_y-x}
    \end{subfigure}~
    \begin{subfigure}{0.24\linewidth}
        \centering
        \includegraphics[page=8,trim=28cm 16cm 28cm 16cm, clip,width=\linewidth]{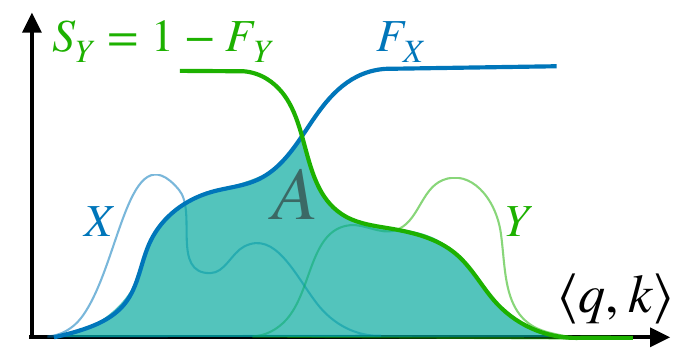}
        \caption{$\mathbb{E}[\max(Y-X, 0)]\!\leq \!A$}
        \label{fig:ub_rc_y-x}
    \end{subfigure}
    \caption{\label{fig:overlap_areas}Illustrations of upper and lower bounds (\cref{thm:oa}) for two types of RC: $\max(X-Y,0)$ and $\max(Y-X,0)$.
    }
\end{figure*}

\subsection{Probabilistic Formalism of Contextualization}\label{sec:formalism}

Let $V$ be a vocabulary of tokens and $\Omega \subset V^*$ the sample space of all finite token sequences. Let $(\Omega, 2^\Omega, P)$ be a probability space, where the probability measure $P(s)$ of a token sequence $s=(t_1,\dots,t_n)\in\Omega$ is defined by an auto-regressive pre-trained transformer. For a sequence $s$, let the attention logit be the function $f_s^{l,h}:[n]\times[n]\to\mathbb{R}$, that maps a pair of tokens at positions $i,j$ to
\begin{align}
    f_s^{l,h}(i,j) = 
    \begin{cases}
        \langle q_j^{l,h},k_i^{l,h}\rangle &\text{ for } i\leq j\leq n \\
        -\infty &\text{ otherwise,}
    \end{cases}
\end{align}
where $q_j^{l,h}$ and $k_i^{l,h}$ are the query and key vectors of $h^\text{th}$ head in $l^\text{th}$ layer. Our framework develops characteristics for individual heads, but we drop the superscript~$^{l,h}$ for brevity. 

A sequence has two parts, $s=p \, \oplus\, g$, where $\oplus$ denotes concatenation, $p=\{t_1, \cdots, t_m\}$ is the sequence of prompt tokens and $g=\{t_{m+1}, \cdots, t_n\}$ is the sequence of generated tokens. The demarcation between $p$ and $g$ need not be rigidly associated with their names, prompt, and generation. 
The $p$ could be a user-given prompt, a conversation history between the user and the language model, or a text that the language model is expected to continue. Similarly, $g$ could be LLM-generated tokens or a portion of an existing text that needs to be analyzed with respect to its previous text $p$. If there is a user question between $p$ and $g$, one may choose to place it at the end of $p$ or the start of $g$. 

Next, we define four random variables (RVs), all illustrated in \cref{fig:illustration}, that capture the notion of contextualization at an attention-head level, a sequence level, and a sub-sequence level. These definitions assume a known strategy to demarcate a token sequence $s$ into $p$ and $g$, i.e., $s=p\oplus g$.

\begin{definition}[Cross-Contextualization]\label{def:cc}
   We define cross-contextualization of an attention head 
   as
\begin{align}\label{eq:cc}
    F_X(X \le x) := \sum_{s \in \Omega} P(s)  \!\!\sum_{\substack{t_i \in p,\, t_j \in g}}\!\! \frac{\mathbf{1}\!\left(f_s(i,j) \le x\right)}{|p||g|},
    \tag{CC}
\end{align}
\end{definition} 
where $\mathbf{1}(\cdot)$ is the indicator function. The \ref{eq:cc} random variable captures the notion of contextualization between the prompt sequence and the generated sequence at a head-level. For a given sequence $s$, it becomes a conditional RV $X_s := X|s$, whose Cumulative Distribution Function (CDF) does not include the outermost summation of \ref{eq:cc}.

Similarly, the next definition captures the contextualization within the generated tokens at a head-level.

\begin{definition}[Self-Contextualization]\label{def:sc}
    We define self-contextualization of an attention head 
    as
    \begin{align}\label{eq:sc}
    \!\!F_Y(Y\! \leq \!y)\! =\! \sum_{s\in \Omega} P(s) \!\! \sum_{t_i,t_j \in g}\!\! \frac{ \mathbf{1}(i\! \leq\! j \, \wedge\, f_s(i,j)\!\leq\! y)}{((|g|+1)|g|)/2} . 
        \tag{SC}
    \end{align}
\end{definition}
For a given sequence $s$, we denote $Y_s=Y|s$ Similar to $X_s$.
Beyond CC and SC, we define their conditional counterparts in \S \ref{sec:ccc_csc} that restrict attention to subsequences of prompt and generated tokens. For $p_1 \subset p$ and $g' \subset g$, conditional CC denoted as $X_{s}(p_1\! \leadsto\! g')$ in \eqref{eq:ccc} captures the influence of $p_1$ on $g'$, while for $g_1, g' \subset g$, conditional SC denoted as $Y_s(g_1\!\leadsto\!g')$ in \eqref{eq:csc} measures the influence of $g_1$ on $g'$.


\subsection{Relative Contextualization (RC)}
\label{sec:rc}

Due to position embeddings and the auto-regressive nature of generation, self-contextualization values are generally higher than cross-contextualization. Yet some influential prompt tokens also receive higher softmax attention, implying their $\langle q,k\rangle$ values exceed those of certain generated tokens, since softmax is order-preserving. This motivates a third type of variable, \emph{relative contextualization}, designed to isolate prompt-specific influence by subtracting out the internal generation bias. 
Formally, if $X_s > Y_s$ with high probability, then the (next-layer) value embeddings of the generated tokens $g$ are heavily affected by the prompt $p$. 

\begin{definition}[Relative Contextualization]\label{def:rc}
    Assuming a sequence $s=p\oplus g$ is given, 
    we define the relative contextualization (RC) as
    \begin{align}\label{eq:rc}
        Z_s = \max\left(X_s - Y_s, 0\right). \tag{RC}
    \end{align}
\end{definition} 
\vspace{-0.2em}
Similar to conditional versions of CC and SC, in \S \ref{sec:cond_rc} we define conditional RC \eqref{eq:cond_rc} denoted as $Z_s((p_1, g_1) \leadsto g')$. 
This measures how much more influence $p_1\subset p$ has on the generated tokens $g'\subset g$ than $g_1 \subset g$ does.


Estimating the central statistics of \ref{eq:rc} requires an operation over the joint distribution of $X$ and $Y$. If there are a large number of prompt and generated tokens, then computing any statistics of $Z$ could become prohibitively expensive, as discussed in \S \ref{sec:comp_complexity}. Note, $X_s$ and $Y_s$ are not independent random variables since the key and query vectors of different tokens are intricately dependent on each other.

In the following, we derive a computationally efficient and practical upper bound for expected \ref{eq:rc} (the same for the conditional RC is in \cref{thm:conditional_oa}), without making any distributional assumptions on $X_s$ or $Y_s$. 
We refer to this upper bound as RC score.

\begin{theorem}[Area Under CDFs]\label{thm:oa}
    The expected relative contextualization $Z$ is upper bounded by the overlap area $A$ between, a) the area under the marginal CDF $F_Y$ of self-contextualization $Y$, and b) the area under the marginal survival function $S_X$ of cross-contextualization $X$:
    \begin{align}\label{eq:rc_ub}
        \mathbb{E}[Z_s] \leq A_s := \int_{-\infty}^{\infty}\!\!\min \left(F_{Y_s}(t),\ S_{X_s}(t)\right) dt,
    \end{align}
    where $S_{X_s}(t)=1-F_{X_s}(t)$, and lower bounded by the area $a$, under $F_Y$ but over $F_X$: 
    \begin{align}\label{eq:rc_lb}
    \mathbb{E}[Z_s] \geq a_s := \int_{-\infty}^{\infty} \max(F_{Y_s}(t)-F_{X_s}, 0) dt.
    \end{align}
\end{theorem} 
\vspace{-0.2em}
Our proof of \cref{thm:oa} in \S \ref{sec:proof} is inspired by \citep{Angelis2021marginalCDF,Vallender1974Wasserstein} and uses copula \citep{Durante2010Copula}. Intuitively, the upper bound in \eqref{eq:rc_ub} captures the area under the overlap region between the values of $X$ and $Y$, as illustrated in \cref{fig:ub_rc_x-y}. 

An alternative relative contextualization can be defined as $Z_s' := \max(Y_s - X_s, 0)$ that captures by how much the SC is more than that of the CC. The lower and upper bounds of $Z'_s$ can be formulated similar to $Z$ as in \cref{thm:oa}, and illustrated in \cref{fig:lb_rc_y-x} and \ref{fig:ub_rc_y-x}.

\subsection{Computational Complexity} \label{sec:comp_complexity}
We analyze the complexity of directly computing expected \ref{eq:rc} and computing its upper bound $A_s$ \eqref{eq:rc_ub} for all the prompt and generated tokens and a given attention head. For direct computation, if we make a simplifying assumption that the joint distribution of $X_s$ and $Y_s$ is a uniform distribution over its discrete support, the expected RC can be calculated as
\vspace{-0.3em}
\begin{align}\label{eq:join_rc}
\mathbb{E}[Z_s]= 
   \sum_{t_i \in p}
   \sum_{\substack{t_j,t_k \in g \\ t_l \in g,\ l \ge k}}
   \frac{\max\!\big(f_s(i,j)-f_s(k,l),0\big)}{|p|\,|g|^2(|g|+1)/2}
\end{align}
using $O(|p| |g|^3)$ computations. We give similar analysis for expected conditional RC $\mathbb{E}[Z_s((p_1, g_1) \leadsto g')]$ in \S \ref{sec:cond_rc}. 

A simpler approximation uses conditional expectation by assuming output tokens are independent and uniformly distributed. We formalize this i.i.d. assumption in \S\ref{sec:independence} and \eqref{eq:iid}.
Although the complexity of this i.i.d. approximation is $O(|p||g|^2)$, which is less than \eqref{eq:join_rc}, we observe that it performs poorly in downstream tasks such as KV-compression (see \cref{tab:ver-qmsum} in \S \ref{sec:independence}). Hence, we do not use this approximation. On the other hand, the upper bound $A_s$ is calculated in $\tilde{O}(|p||g| + |g|^2)$ computations, where $\tilde{O}(n)=O(n\log(n))$, with the Lebesgue integration approach in \cref{alg:oa}.

\begin{minipage}[h]{0.4\textwidth}
It involves sorting samples of $X_s$ and $Y_s$, first individually
then jointly, to obtain the unique breakpoints $B$. With appropriate indexing, the complexity of computing CDFs and the minimum values in Lines \ref{algst:Fx}, \ref{algst:Fy}, and \ref{algst:mv} becomes linear with respect to the number of midpoints $M$ in Line \ref{algst:mp}. 
The complexity above corresponds to computing $A_s$ at the head level.  
At the subsequence level, computing $A_s\big((p_1, g_1) \leadsto g\big)$ (\eqref{eq:cond_rc_ub}) has complexity $\tilde{O}(|p_1||g| + |g_1||g|)$.  
\cref{alg:oa} naturally extends to compute the upper~\eqref{eq:rc_ub} and lower~\eqref{eq:rc_lb} bounds for both RC variants in \cref{fig:overlap_areas} by modifying Line~\ref{algst:mv}.
\end{minipage}
\hfill
\begin{minipage}[h]{0.58\textwidth}
\vspace{-3mm}
\begin{algorithm}[H]
\caption{Area under \( \min(F_{Y_s}, 1 - F_{X_s}) \)}\label{alg:oa}
\begin{algorithmic}[1]
\setlength{\itemsep}{0pt}
\setlength{\topsep}{2pt}
\Require Samples of $X_s$ and $Y_s$
\Ensure Overlap area $A_s$
\State $X \gets \mathtt{sort}(X_s), Y\gets\mathtt{sort}(Y_s)$  \label{algst:sort}
\State \( B \gets \texttt{unique}(\texttt{sort}(X \cup Y)) \)\label{algst:sort_unique}
\State \( L \gets B_{1:\ell - 1} \), \( R \gets B_{2:\ell} \) \hfill $\triangleright$ $\ell = |B|$
\State \( M \gets (L + R) / 2 \) \hfill $\triangleright$ Midpoints \label{algst:mp}
\State \( W \gets R - L \) \hfill $\triangleright$ Widths
\For{each midpoint \( m_i \in M \)}
    \State \( F_X(m_i) \gets \frac{1}{n} \sum_j \mathbf{1}[X_j \le m_i] \) \label{algst:Fx}
    \State \( F_Y(m_i) \gets \frac{1}{m} \sum_j \mathbf{1}[Y_j \le m_i] \) \label{algst:Fy}
    \State \( v_i \gets \min(F_Y(m_i), 1 - F_X(m_i)) \)  \label{algst:mv}
\EndFor
\State \( A \gets \sum_i v_i \cdot W_i \quad \)  \hfill $\triangleright$ unique $v_i$s $\Rightarrow$ Lebesgue Integral
\State \Return \( A \)
\end{algorithmic}
\end{algorithm}
\end{minipage}

\section{Applications of Relative Contextualization}


Depending on the choice of prompt and generation subsets $p_1 \subset p$ and $g_1 \subset g$, Equation~\ref{eq:rc} supports different applications. We focus on two use cases.
\vspace{-1em}

\subsection{Post-hoc Attribution to Context Tokens}\label{sec:attribution}
\vspace{-0.5em}
We now consider the problem of post-hoc attribution, where explanations are computed after generation is complete, to link generated content to its supporting context.
Given a generated token span $g' = [t_{j_1}, \ldots, t_{j_2}]$ and a collection of candidate prompt spans $S = \{p_1, \ldots, p_{|S|}\}$, the goal is to identify which spans in $S$ are most responsible for producing $g'$. Each prompt span $p_i$ may correspond to a retrieved chunk in a RAG setting, a few-shot example used for in-context learning, or a singleton prompt token $p_i = \{t_i\}$. Our attribution method proceeds in three steps.


\textbf{Step 1:} Compute the RC score $A_s^h$ \eqref{eq:rc_ub} over the full sequence $s$ for every attention head $h$, and select the top-$k$ heads $\mathcal{H}_k$.

\textbf{Step 2:} For each candidate span $p_i \in S$ and head $h \in \mathcal{H}_k$, compute the conditional RC score $A_s^h((p_i, g')\leadsto g')$ from $p_i$ to the generation span $g'$.

\textbf{Step 3:} Aggregate across heads to obtain a normalized attribution score for each span $p_i$:
\begin{align}\label{eq:att_score}
\mathrm{RC}(p_i)
= \frac{\sum_{h \in \mathcal{H}_k} A_s^h((p_i, g') \leadsto g')}
{\sum_{p_{i'} \in S} \sum_{h \in \mathcal{H}_k} A_s^h((p_{i'}, g') \leadsto g')} .
\end{align}
We attribute $g'$ to the span with the largest $\mathrm{RC}(p_i)$. 




\begin{figure*}[t]
\centering
\begin{subfigure}{0.25\textwidth}
    \centering
    \includegraphics[width=\linewidth]{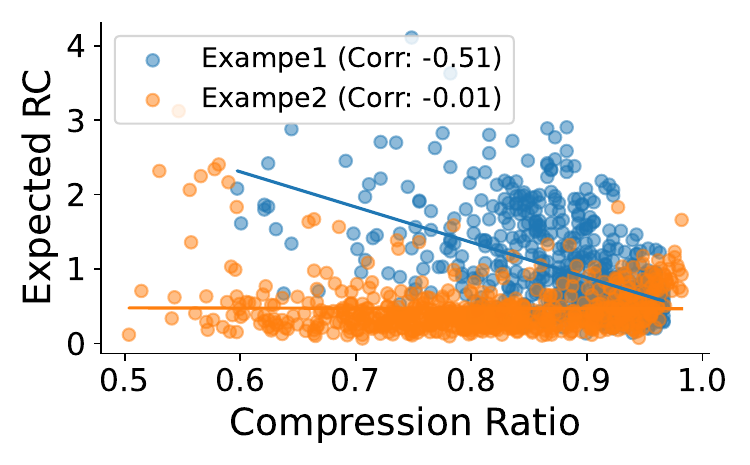}
    \caption{Two examples}
    \label{fig:corr}
\end{subfigure}~
\begin{subfigure}{0.25\textwidth}
    \centering
    \includegraphics[width=\linewidth]{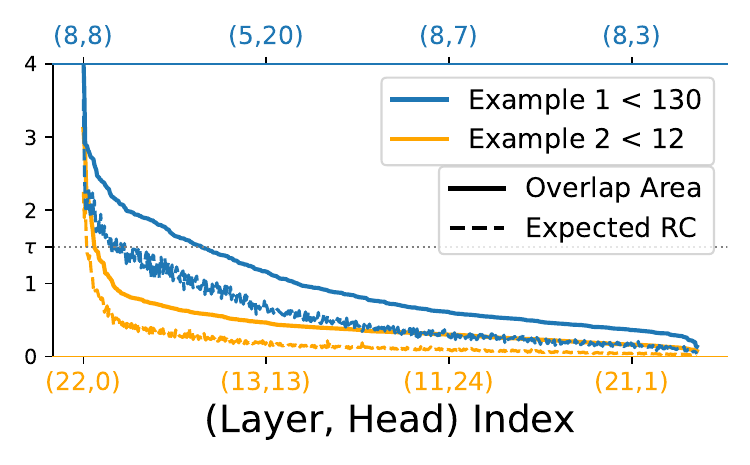}
    \caption{Two examples}
    \label{fig:context_compare}
\end{subfigure}~
\begin{subfigure}{0.25\textwidth}
    \centering
    \includegraphics[width=\linewidth]{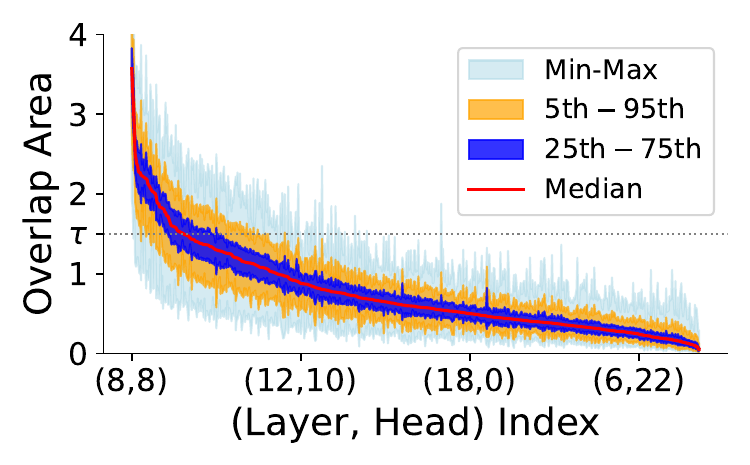}
    \caption{QMSum dataset}
    \label{fig:qmsum_erc}
\end{subfigure}~
\begin{subfigure}{0.25\textwidth}
    \centering
    \includegraphics[width=\linewidth]{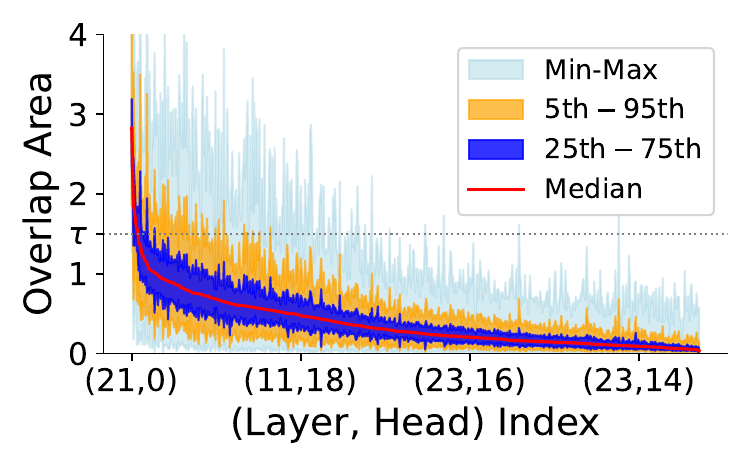}
    \caption{SQuAD dataset}
    \label{fig:squad_erc}
\end{subfigure}

\caption{\label{fig:head_roles} Using LLaMA-3B (28 layers, 24 heads) on the listed datasets. \cref{fig:context_compare} shows $\mathbb{E}[RC]$~\eqref{eq:join_rc} and its upper bound~\eqref{eq:rc_ub} (overlap area in \cref{fig:ub_rc_x-y}) for all heads. \cref{fig:qmsum_erc} and \ref{fig:squad_erc} show head RC scores across datasets, sorted by median. \cref{fig:corr} shows the anti-correlation between head compression ratio and RC score (each point is a head).}


\vspace{-4mm}
\end{figure*}
\vspace{-4mm}
\subsection{KV Cache Compression}\label{sec:kv_comp}
\vspace{-0.5em}
In attention head, the value vector $v_j \in \mathbb{R}^d$ for token $t_j$ is computed with and without KV eviction as
\begin{align}\label{eq:ve}
v_j^* = \frac{\sum_{i\in[j]} e^{f(i,j)} v_i}{\sum_{i\in[j]} e^{f(i,j)}},
\quad
\hat{v}_j = \frac{\sum_{i\in [j]\setminus p_e} e^{f(i,j)} v_i}{\sum_{i\in [j]\setminus p_e} e^{f(i,j)}},
\end{align}
where $p_e \subset [|p|]$ is the index set of token-positions with evicted KV entries. An ideal eviction policy minimizes the degradation in value fidelity over generated tokens $g$:
\begin{align}\label{eq:kv_comp_metric}
p_e^* = \arg\min_{p_e \subseteq p} \frac{1}{|g|} \sum_{t_j \in g} \| v_j^* - \hat{v}_j \|_2 ,
\end{align}
which is combinatorial and infeasible during decoding since $g$ is unknown. 
Following SnapKV~\citep{li2024snapkv}, we approximate $g$ using a small window of recent prompt tokens $\hat{g}\subset p$. Treating expected RC as a measure of contextual significance, we evict the KV pair of a token $t_i$ if its singleton contribution is small relative to rest of the prompt:
\begin{align}\label{eq:eviction}
A_p((p_i, \hat{g}) \leadsto \hat{g})
\leq
c \, A_p((p \setminus \hat{g}, \hat{g}) \leadsto \hat{g}),
\end{align}
where $p_i=\{t_i\}$, $A_p$ is defined analogously to condition RC score \ref{eq:cond_rc_ub} over the prompt $p$ (here $s=p=(p\setminus\hat{g}) \oplus \hat{g}$) and $c$ controls the compression rate. Eviction is adaptive across heads: for a fixed $c$, each head selects a different $p_e$, reflecting differences in contextual load. 

\section{Experiments}
\label{sec:eval}

We first study the behavior of \textsc{RC} by analyzing its expected scores and upper bounds across attention heads, and then evaluate on two tasks: (i) attribution, (ii) KV-cache eviction. 


\vspace{-0.2em}
\textbf{Datasets and LMs:} For head analysis and KV-cache eviction, we use 3 benchmarks: $2000$ SQuAD v2.0~\citep{rajpurkar2018know} (span QA with unanswerable queries), $200$ QMSum~\citep{zhong-etal-2021-qmsum} (query-focused meeting summarization), and $2000$ 2WikiMultiHop~\citep{ho-etal-2020-constructing} (multi-hop RC over linked Wikipedia). For attribution, we use $1300$ QuoteSum~\citep{schuster2023semqa} (summaries with annotated source spans) and $200$ \textsc{Veri-Gran}~\citep{phukan2024peering} (grounded-generation attribution). We evaluate 3 LMs from two families: LLaMA-3.2-3B and LLaMA-3.1-8B Instruct (LLaMA; \citealp{grattafiori2024llama}), and Qwen3-8B (Qwen; \citealp{yang2025qwen3technicalreport}). (Ref.~\S \ref{sec:appendix_reproducibility} for details).

\vspace{-0.2em}
\subsection{Roles of Attention Heads Across Tasks}\label{sec:head_role}



\vspace{-0.1em}
We analyze if Relative Contextualization (RC) can distinguish relevant vs. irrelevant context, contrast complex summarization from simple QA, and how contextualizing heads varies with task difficulty.

\vspace{-0.2em}
\textbf{Two Contrasting Examples:} We provide two  examples in \S \ref{sec:appendix_prompt_examples}, each with a prompt as the context and the generation as a question-answer pair. In Example 1, the context supports the question; in Example 2, it does not. \cref{fig:context_compare} shows that head-level RC scores, i.e., the relative contextualization (overlap) between prompt and generation, are uniformly lower for Example 2 with the irrelevant context across all heads and layers. This is reflected in the count of heads whose overlap upper bound exceeds a threshold \(\tau=1.5\); only 12 heads are responsible for contextualization for Example 2, compared to 130 heads for Example 1.


\vspace{-0.2em}
\textbf{Sensitivity towards Task Complexity:}  Comparing Figures \ref{fig:qmsum_erc} and \ref{fig:squad_erc}, we observe that the RC scores are substantially higher for QMSum, complex multi-sentence summarization tasks, than for SQuAD, simple single-hop QA tasks. On average, QMSum requires $9\times$ more heads ($p<0.05$) than SQuAD, with RC scores higher than $\tau$, 
indicating that complex tasks recruit a broader ensemble of attention heads.
We analyze this phenomenon further in \S \ref{sec:task_complexity}.



\vspace{-0.2em}
\textbf{Consistency of Heads Across Examples:} 
Figures \ref{fig:qmsum_erc} and \ref{fig:squad_erc} also presents the distribution of overlap‐area (min–max, 5th–95th, 25th–75th, median) for all 24×28 head-layer pairs on QMSum and SQuAD datasets respective, revealing a pronounced long‐tail behavior: a handful of heads exhibit persistently large overlap across inputs, while most heads remain low. 
In \S \ref{sec:RC_distribution}, we provide detailed head-wise RC distribution plots and heat maps for each dataset and model.

\begin{table}[t]
  \begin{minipage}[t]{0.49\linewidth}
    \caption{Chunk-level accuracy for attributing extractive spans on
    QuoteSum and \textsc{Veri-Gran} (“L3.1” = LLaMa-3.1, “L3.2” = LLaMa-3.2).
    For LLaMa-3.2-3B and Qwen3-8B, values are pre-softmax.
    “HS” = best single hidden-state layer.}
    \label{tab:table_attribution}
    \centering
    \resizebox{\linewidth}{!}{%
      \begin{tabular}{@{}lcc@{}}
      \toprule
      Model & QuoteSum & \textsc{Veri-Gran} \\
      \midrule
      GPT-3.5 (inline)           & 90.18 & 26.40 \\
      GPT-4 (inline)             & 90.59 & 62.11 \\
      BM25                       & 75.72 & 68.20 \\
      GTR                        & 72.57 & 53.15 \\
      MonoT5                     & 89.24 & 67.43 \\
      L3.1-8B (HS~\citep{phukan2024peering}) & 86.10 & 71.90 \\
      L3.2-3B (HS; layers 16/14/25) & 86.80 & 71.60 \\
      Qwen3-8B (HS; layers 16/14/25) & 79.60 & 63.40 \\
      \midrule
      \arrayrulecolor{black!30}\specialrule{.5pt}{2pt}{2pt}
      L3.1-8B (all heads)        & 90.54 & 77.91 \\
      L3.1-8B (least RC, $k{=}20$) & 29.49 & 2.81 \\
      \rowcolor{mygreen} L3.1-8B (\textbf{most RC}, $k{=}20$) & \textbf{93.91} & \textbf{79.37} \\
      \specialrule{.5pt}{2pt}{2pt}\arrayrulecolor{black}
      \arrayrulecolor{black!30}\specialrule{.5pt}{2pt}{2pt}
      L3.2-3B (all heads)        & 92.20 & 78.70 \\
      L3.2-3B (least RC, $k{=}20$) & 31.80 & 1.90 \\
      \rowcolor{mygreen} L3.2-3B (\textbf{most RC, $k{=}20$}) & \textbf{92.70} & \textbf{79.00} \\
      \specialrule{.5pt}{2pt}{2pt}\arrayrulecolor{black}
      \arrayrulecolor{black!30}\specialrule{.5pt}{2pt}{2pt}
      Qwen3-8B (all heads)       & 39.10 & 5.60 \\
      Qwen3-8B (least RC, $k{=}20$) & 31.90 & 4.40 \\
      \rowcolor{mygreen} Qwen3-8B (\textbf{most RC, $k{=}20$}) & \textbf{92.80} & \textbf{79.40} \\
      \specialrule{.5pt}{2pt}{0pt}\arrayrulecolor{black}
      \bottomrule
      \end{tabular}
    }%
  \end{minipage}%
  \hfill
  \begin{minipage}[t]{0.49\linewidth}
    \centering
    \phantomsection\label{fig:llama_head}%
    \includegraphics[width=\linewidth]{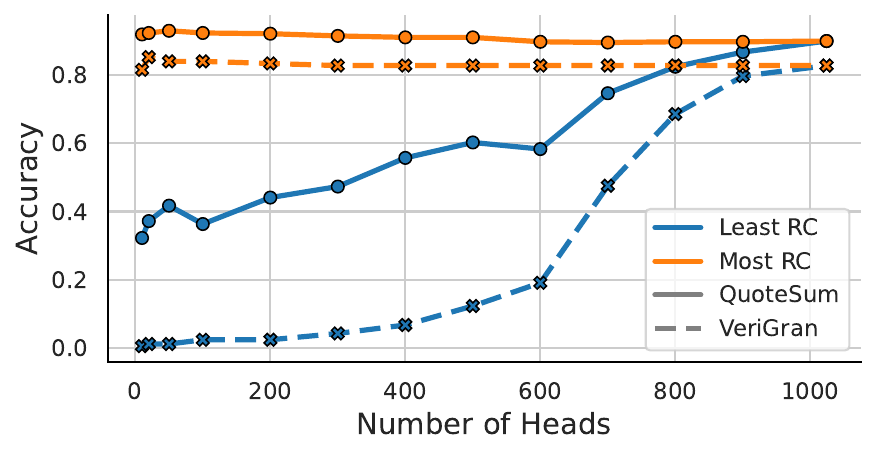}\\[-2pt]
    {\small (a) LLaMa-3.1-8B}\\[2pt]
    \phantomsection\label{fig:qwen_head_robust}%
    \includegraphics[width=\linewidth]{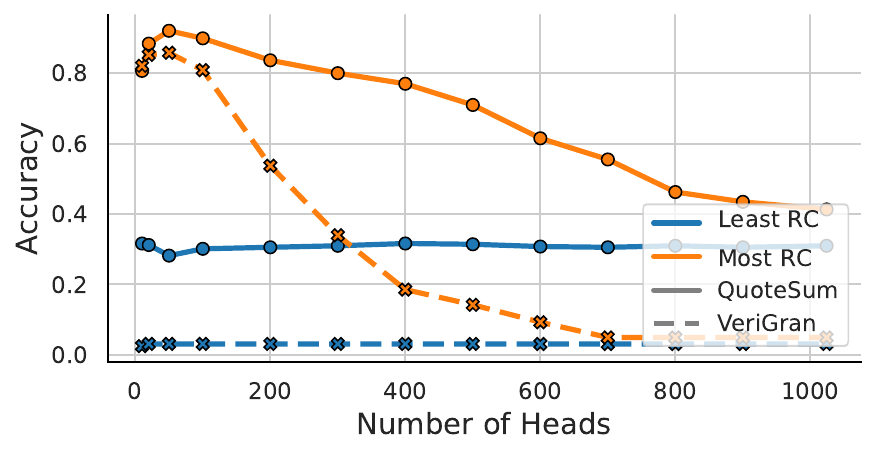}\\[-2pt]
    {\small (b) Qwen3-8B}\\[4pt]
    \captionof{figure}{Attribution accuracy vs.\ \# heads.
    \emph{Most RC}: descending RC; \emph{Least RC}: ascending.}
    \label{fig:ablation_top_k_bottom_k}
  \end{minipage}
  \vspace{-1em}
\end{table}

\vspace{-1.0em}
\subsection{Chunk-Level Post-hoc Context Attribution}\label{sec:exp_attribution}
\vspace{-0.2em}
\vspace{-0.2em}
\textbf{Baselines:} We focus on training-free, inference-only attribution methods that are computationally efficient for deployment in real production systems. To the best of our knowledge, the zero-train methods include the hidden state (HS) approach~\citep{phukan2024peering} and retrieval-based methods, such as BM25 (sparse)~\citep{BM25}, GTR (dense)~\citep{GTR}, and MonoT5~\citep{MTR}. We also compare against GPT based attribution as done by \cite{phukan2024peering}.



\vspace{-0.2em}
\textbf{Quantitative Results:} 
We report results for LLaMA3.1-8B in \cref{tab:table_attribution}, 
On QuoteSum, using the top-$k$ heads improves attribution accuracy by 3\% over using all heads, while bottom-$k$ causes a 61\% drop. On \textsc{Veri-Gran}, top-$k$ heads provide a 2\% gain, whereas bottom-$k$ heads fall to 2.8\% (near random) for LLaMa-3.1-8B. These results confirm the importance of heads with high RC scores computed from pre-softmax attention logits. Interestingly, selecting heads based on post-softmax RC scores performs worse than all heads (ref. \cref{tab:table_attribution_pre_post} in \S \ref{sec:post_vs_pre}).
We further demonstrate the impact and genaralizability of our technique in identifying top-K RC heads in Table~\ref{tab:table_attribution} with two more models - a smaller \textbf{LLaMA-3.2-3B} and \textbf{Qwen3-8B} with different architecture.
Attribution using top-k RC heads consistently outperforms baselines, achieving 5.9–13.2\% gains over HS on QuoteSum and 7.4–16.0\% gains on Veri-Gran. Least-RC heads perform near chance. This effect is particularly pronounced in Qwen3-8B, where attribution drops from 92.8\% (most-RC) to 39.1\% (all heads), approaching the 31.9\% of least-RC heads. LLaMA models show smaller drops (e.g., 93.9→90.5).  


\vspace{-0.2em}
\textbf{Head-Selection Robustness:}  
To dive deeper, \cref{fig:ablation_top_k_bottom_k} shows attribution accuracy as a function of $k$ for LLama-3.1-8B in (a) and Qwen3-8B in (b). 
When using the top‐$k$ heads, accuracy remains near its peak even for small $k$. In contrast, accuracy plummets as more bottom‐$k$ heads are included, even up to 800 (out of 1024) heads. This behavior is even more prominent in the Qwen3-8B model as shown in \cref{fig:qwen_head_robust}. 
This shows selecting heads beyond the optimal set identified by our method can significantly degrade accuracy. We provide qualitative RC-based attribution examples in \S\ref{sec:attribution_qualitative}.



\vspace{-1em}

\subsection{KV Cache Eviction}\label{sec:kv_comp_exp}
\vspace{-0.5em}


    

\textbf{Baselines:}
We compare our approach, \textsc{RCStat}, against state-of-the-art methods: Knorm~\citep{liu2024minicache}, SnapKV~\citep{li2024snapkv}, StreamingLLM~\citep{xiao2023efficient}, and TOVA~\citep{liu2023scissorhands}. Notably, both SnapKV and TOVA operate on post-softmax attention scores.

\begin{figure*}[t]
    \centering
    \begin{subfigure}{\textwidth}
    \centering        \includegraphics[width=0.8\linewidth]{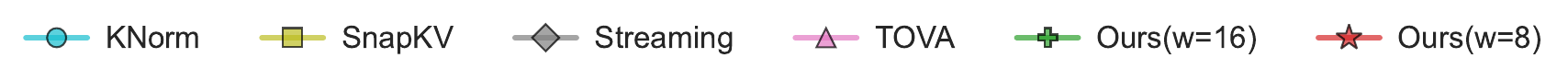}
    \end{subfigure}\\[-2mm]
    \begin{subfigure}{0.25\textwidth}
        \includegraphics[width=\linewidth]{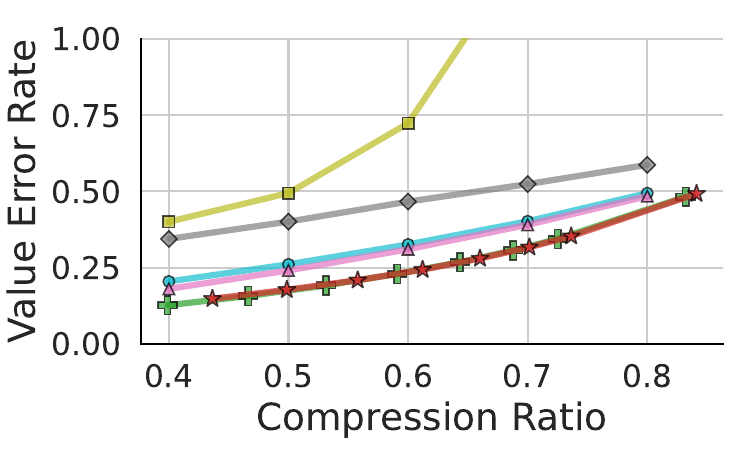}
        \caption{3B model}
        \label{fig:dv_qmsum_3b}
    \end{subfigure}~
    \begin{subfigure}{0.25\textwidth}
        \includegraphics[width=\linewidth]{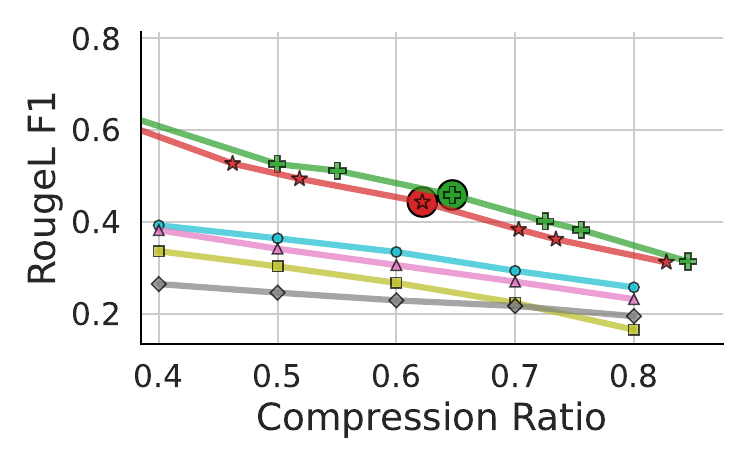}
        \caption{3B model}
        \label{fig:rLf1_qumsum_3b}
    \end{subfigure}~
    \begin{subfigure}{0.25\textwidth}
        \includegraphics[width=\linewidth]{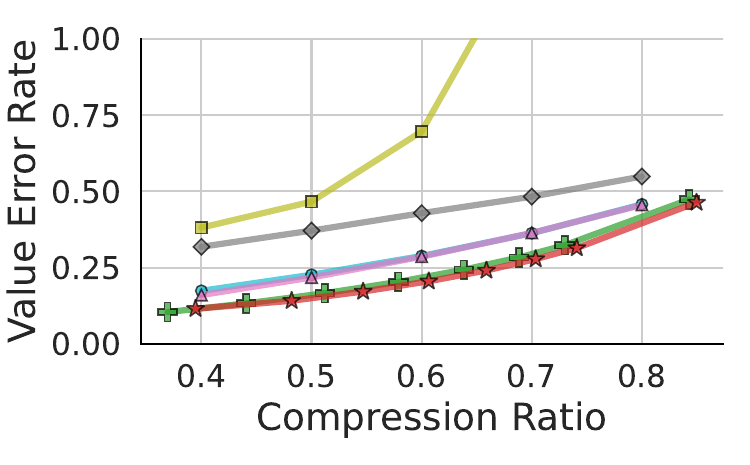}
        \caption{8B model}
        \label{fig:dv_qmsum_8B}
    \end{subfigure}~
    \begin{subfigure}{0.25\textwidth}
        \includegraphics[width=\linewidth]{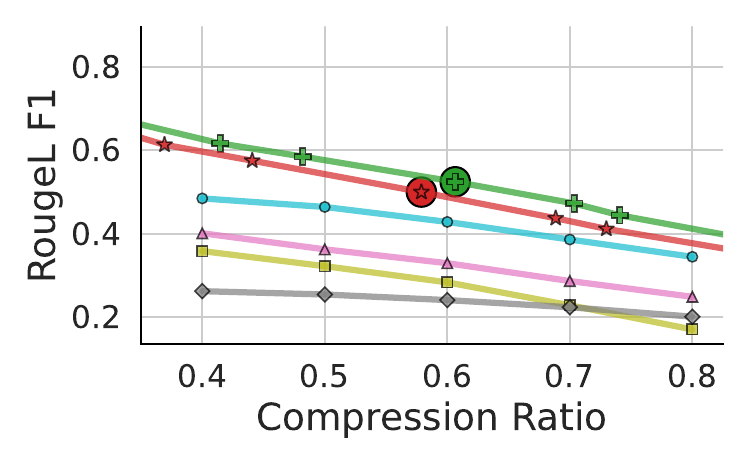}
        \caption{8B model}
        \label{fig:rLf1_qmsum_8B}
    \end{subfigure}~
    \vspace{-1mm}  
    \caption{\label{fig:kv_comp_qmsum} KV-cache eviction performance on QMSum using LLaMA-3.2-3B and LLaMA-3.1-8B. We report Value Error Rate (VER↓) and generation quality (RL-F1↑) across different strategies.}
    \vspace{-5mm}
\end{figure*}
\begin{figure*}[t]
    \centering
    \begin{subfigure}{\textwidth}
    \centering
        \includegraphics[width=0.8\linewidth]{Figures/only_legend.pdf}
        \label{fig:leg}
    \end{subfigure}\\[-2mm]
    \begin{subfigure}{0.25\textwidth}
        \includegraphics[width=\linewidth]{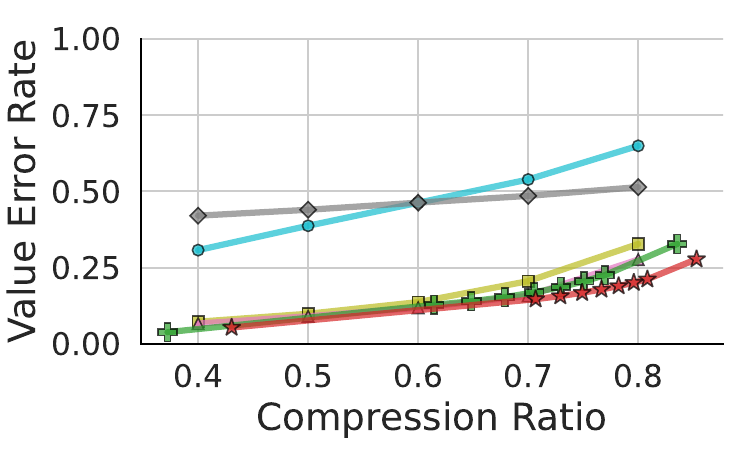}
        \caption{2WikiMultiHop}
        \label{fig:dv_2wiki_8B}
    \end{subfigure}~
    \begin{subfigure}{0.25\textwidth}
        \includegraphics[width=\linewidth]{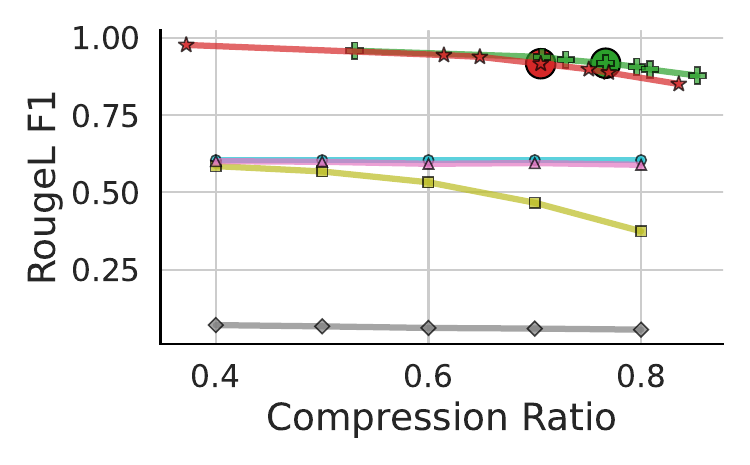}
        \caption{2WikiMultiHop}
        \label{fig:rLf1_2wiki_8B}
    \end{subfigure}~
    \begin{subfigure}{0.25\textwidth}
        \includegraphics[width=\linewidth]{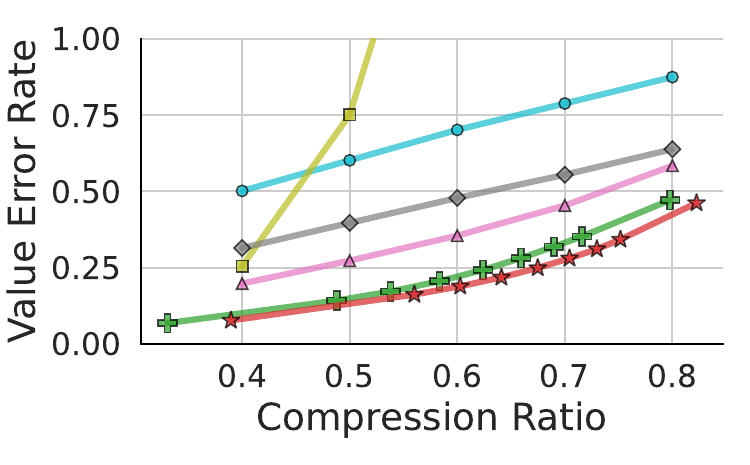}
        \caption{SQuAD}
        \label{fig:dv_squad_8B}
    \end{subfigure}~
    \begin{subfigure}{0.25\textwidth}
        \includegraphics[width=\linewidth]{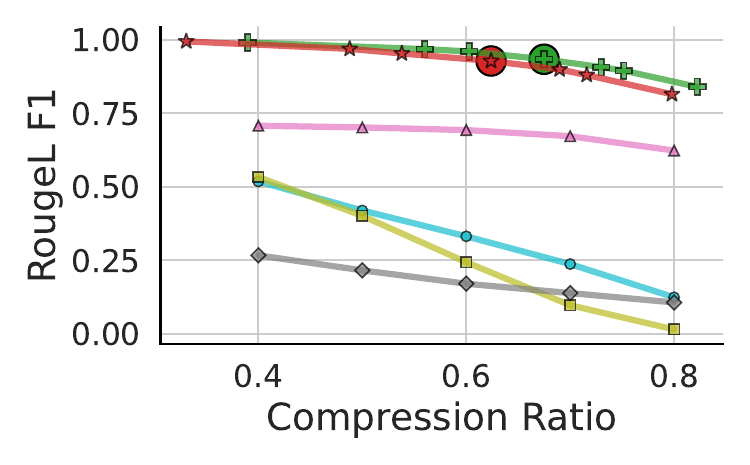}
        \caption{SQuAD}
        \label{fig:rLf1_squad_8B}
    \end{subfigure}~
    \vspace{-1mm}  
    \caption{\label{fig:kv_comp_2wiki_squad} KV-cache eviction results on 2WikiMultiHop and SQuAD v2.0 using LLaMA-3.1-8B.
    }
\end{figure*}

\textbf{Metrics:} The primary metric is the overall compression ratio: the ratio between the evicted KV-cache size and total KV-cache size across all layers and heads. We evaluate the performance using ROUGE-1/L~\citep{lin2004rouge} and Value Error Rate (VER), the objective in Eq.~\eqref{eq:kv_comp_metric} normalized by $v_j^*$ inside the summation. For VER, we use the last layer's value vectors of LLM and average across all heads and samples in the dataset. 
In \S \ref{sec:decoding_strategy}, we discuss how VER is more robust to variations in decoding strategies compared to ROUGE, making it a suitable metric for KV-compression. 

\textbf{Memory and Computational Efficiency:} 
We discuss computational efficiency in \S \ref{sec:efficiency_kvc}. RCStat achieves up to 32\% higher memory savings than the strongest baseline at the same accuracy. Prefill dominates runtime (~3 ms per layer on Llama-3.2-3B with ~1000 tokens), while RCStat adds only 0.68 ms per layer for eviction, comparable to SNAP (0.66 ms) and TOVA (0.65 ms). Decoding latency then improves with compression ratio, while preserving answer quality under aggressive compression.


\textbf{Experimental Setup:} We build upon the baseline implementations provided in the KVPress package~\citep{jegou2024kvpress}, using default values for all method-specific hyperparameters. All baselines are evaluated at compression ratios of $0.4, 0.5, \cdots, 0.8$. For \textsc{RCStat}, the compression ratio is controlled via the parameter $c$ in \eqref{eq:eviction}, where larger values of $c$ result in more aggressive eviction and thus higher compression. We vary $c$ over ${0.2, 0.7, 0.8, 1.0, 1.2, 1.3, 1.8}$, with the default setting $c{=}1$ highlighted using a circle in all plots. Here, we evaluate our method with window sizes of the last few tokens (see \S \ref{sec:kv_comp}), with $w \in {8, 16}$.  \S \ref{sec:window_size} shows results of further window size variation.

\textbf{Generation-Compression Tradeoff:}  
We present some of our results in Figures \ref{fig:kv_comp_qmsum} and \ref{fig:kv_comp_2wiki_squad}. All results for LLaMA models are in \S \ref{sec:llama_results_kvc} and for Qwen model in \S \ref{sec:qwen_results_kvc}.
We show the results for VER metric in Figures \ref{fig:dv_qmsum_3b}, \ref{fig:dv_qmsum_8B},
\ref{fig:dv_2wiki_8B} and \ref{fig:dv_squad_8B}. Across all datasets and models, our method incurs the least VER for all compression ratios. These results show that the fidelity of the internal representation of generated tokens is best preserved in our method. Similarly, results in Figures  \ref{fig:rLf1_qumsum_3b}, \ref{fig:rLf1_qmsum_8B}, \ref{fig:rLf1_2wiki_8B}, and \ref{fig:rLf1_squad_8B}, our method achieves the best solution frontier in the trade-off between compression ratio and RougeL F1 score. 
We notice that, although a lower VER implies higher RougeL, its inverse is not necessarily true: the ordering of solution frontiers of baselines for VER is not the same in RougeL-F1. This is expected, since RougeL measures n-gram output text overlap. In fact, even at $80\sim90\%$ compression the an LLM generates answers, not from grounding in the context, but internal model weights learnt during pre-training \citep{chuang2024lookback, feldman2023trapping, xu2024hallucination}. 


\textbf{Adaptive Head‐wise Eviction:} 
Our method adaptively evicts KV-caches at the level of individual attention heads. To analyze this, we examine the correlation between head-wise compression ratios and RC scores in \cref{fig:corr}, using the same examples as \S \ref{sec:head_role}. We report only means; variances and other statistics are provided in \S \ref{sec:supp_material}. As expected, heads with higher RC scores face less eviction, reflecting their importance for context-grounded generation. In other words, such heads are highly informative and thus less compressible. We leverage these informative heads for attribution.

\vspace{-1em}

\section{Conclusion}
\label{sec:conclusion}
\vspace{-0.5em}
We present \textsc{RCStat}, a statistical framework that computes Relative Contextualization (RC) from pre-softmax attention logits with an efficient upper bound, enabling practical KV-cache compression and attribution without retraining; empirically, it achieves state-of-the-art accuracy on both while reducing compute. Analysis shows (1) RC quantifies an LLM’s contextualization effort (task difficulty), (2) attention heads exhibit stable, example-agnostic specialization, and (3) the most influential heads cluster in middle layers, consistent with prior work. Explaining \textit{why} and \textit{how} these patterns arise remains future work.

\bibliographystyle{plainnat}
\bibliography{references}

\newpage
\section*{NeurIPS Paper Checklist}

\begin{enumerate}

\item {\bf Claims}
    \item[] Question: Do the main claims made in the abstract and introduction accurately reflect the paper's contributions and scope?
    \item[] Answer: \answerYes{}
    \item[] Justification: The abstract and introduction clearly state all three contributions: (1) the RCStat statistical framework based on pre-softmax logit distributions, (2) the RC-based attribution method, and (3) the KV-cache compression application. Theoretical results (Theorem~3.4) and experimental scope (three models, five benchmarks) are faithfully described.

\item {\bf Limitations}
    \item[] Question: Does the paper discuss the limitations of the work performed by the authors?
    \item[] Answer: \answerYes{}
    \item[] Justification: Limitations are discussed in the Impact Statement and the Conclusion. Specifically, we note that the independence approximation of generated tokens reduces accuracy in some settings (\S\ref{sec:independence}), that scalability at token-level attribution is challenging due to the cost of computing fine-grained overlap areas, and that experiments are limited to models up to 8B parameters.

\item {\bf Theory assumptions and proofs}
    \item[] Question: For each theoretical result, does the paper provide the full set of assumptions and a complete (and correct) proof?
    \item[] Answer: \answerYes{}
    \item[] Justification: All theorems (including Theorem~3.4 on the upper bound via the Fréchet--Hoeffding inequality) state their assumptions explicitly and have complete proofs in Appendix~\ref{sec:ccc_csc}.

\item {\bf Experimental result reproducibility}
    \item[] Question: Does the paper fully disclose all the information needed to reproduce the main experimental results of the paper to the extent that it affects the main claims and/or conclusions of the paper (regardless of whether the code and data are provided or not)?
    \item[] Answer: \answerYes{}
    \item[] Justification: Appendix~\ref{sec:appendix_reproducibility} (Reproducibility Statement) lists all models with versions and licenses, all datasets with versions and URLs, hyperparameters (compression ratios, window sizes, top-$k$), software environment (PyTorch 2.1, CUDA 12.1, A100 80GB), and a link to the anonymized code repository.

\item {\bf Open access to data and code}
    \item[] Question: Does the paper provide open access to the data and code, with sufficient instructions to faithfully reproduce the main experimental results, as described in supplemental material?
    \item[] Answer: \answerYes{}
    \item[] Justification: An anonymized code repository is provided at \url{https://anonymous.4open.science/r/RCStat-8E3D/README.md}. All datasets and models used are publicly available with documented licenses (see Appendix~\ref{sec:appendix_reproducibility}).

\item {\bf Experimental setting/details}
    \item[] Question: Does the paper specify all the training and test details (e.g., data splits, hyperparameters, how they were chosen, type of optimizer) necessary to understand the results?
    \item[] Answer: \answerYes{}
    \item[] Justification: RCStat is training-free, so there is no optimizer or training split. Inference hyperparameters (compression ratios $\in \{0.4, 0.5, 0.6, 0.7, 0.8\}$, window sizes $w \in \{1,2,4,8,16,32,64\}$, top-$k$ heads $k=20$) are reported in \S\ref{sec:experiments} and Appendix~\ref{sec:appendix_reproducibility}.

\item {\bf Experiment statistical significance}
    \item[] Question: Does the paper report error bars suitably and correctly defined or other appropriate information about the statistical significance of the experiments?
    \item[] Answer: \answerYes{}
    \item[] Justification: While we report the mean values in the plots in the main paper to keep the plots less cluttered, in the \textit{Supplementary}, we provide tables reporting the variability (standard deviation numbers) corresponding to the plots presented in the main paper. 

\item {\bf Experiments compute resources}
    \item[] Question: For each experiment, does the paper provide sufficient information on the computer resources (type of compute workers, memory, time of execution) needed to reproduce the experiments?
    \item[] Answer: \answerYes{}
    \item[] Justification: Appendix~\ref{sec:appendix_reproducibility} states that experiments were run on a single A100 80GB GPU with PyTorch 2.1 and CUDA 12.1. Runtime analysis and efficiency comparisons are provided in Appendix~\ref{sec:kv_efficiency}.

\item {\bf Code of ethics}
    \item[] Question: Does the research conducted in the paper conform, in every respect, with the NeurIPS Code of Ethics \url{https://neurips.cc/public/EthicsGuidelines}?
    \item[] Answer: \answerYes{}
    \item[] Justification: This work analyzes attention patterns in publicly available LLMs using publicly available datasets. No new data was collected from human subjects and no models were deployed in production. The work conforms with the NeurIPS Code of Ethics.

\item {\bf Broader impacts}
    \item[] Question: Does the paper discuss both potential positive societal impacts and negative societal impacts of the work performed?
    \item[] Answer: \answerYes{}
    \item[] Justification: \textsc{RCStat} is a statistical framework designed to improve the efficiency and interpretability of large language models by quantifying contextual influence through pre-softmax attention signals.
     
    \emph{Positive impacts:} The attribution capabilities provide a training-free method for practitioners to audit and understand model behavior, enabling users to verify whether LLM answers are actually backed by input context. By enabling more principled KV-cache compression, our work reduces the memory and computational requirements of LLM inference, lowering the environmental footprint of large-scale deployment and increasing accessibility in resource-constrained environments.

    \emph{Limitations and negative impacts:} We emphasize that these attribution signals reflect internal statistical correlations and should not be interpreted as definitive explanations of a model's underlying reasoning. We do not foresee significant negative societal impacts specific to this work beyond those already associated with LLM deployment in general; the techniques are intended as analytical and efficiency tools whose broader ethical implications are aligned with established considerations in responsible ML development.

\item {\bf Safeguards}
    \item[] Question: Does the paper describe safeguards that have been put in place for responsible release of data or models that have a high risk for misuse (e.g., pre-trained language models, image generators, or scraped datasets)?
    \item[] Answer: \answerNA{}
    \item[] Justification: This paper does not release new pre-trained models or scraped datasets. We release only analysis code that operates on existing publicly available models (Llama-3, Qwen3), which already have their own usage licenses.

\item {\bf Licenses for existing assets}
    \item[] Question: Are the creators or original owners of assets (e.g., code, data, models), used in the paper, properly credited and are the license and terms of use explicitly mentioned and properly respected?
    \item[] Answer: \answerYes{}
    \item[] Justification: Appendix~\ref{sec:appendix_reproducibility} lists every asset used (KVPress codebase, Llama-3.1-8B, Llama-3.2-3B, Qwen3-8B, QMSum, 2WikiMultihopQA, SQuAD v2.0, QuoteSum, Verifiability-Granular) with corresponding citations, URLs, access dates, and license names (Apache 2.0, Llama 3.1/3.2 Community License, MIT, CC BY-SA 4.0).

\item {\bf New assets}
    \item[] Question: Are new assets introduced in the paper well documented and is the documentation provided alongside the assets?
    \item[] Answer: \answerNA{}
    \item[] Justification: No new datasets or models are introduced. The only new artifact is the RCStat analysis code, provided anonymously at submission time with a README.

\item {\bf Crowdsourcing and research with human subjects}
    \item[] Question: For crowdsourcing experiments and research with human subjects, does the paper include the full text of instructions given to participants and screenshots, if applicable, as well as details about compensation (if any)?
    \item[] Answer: \answerNA{}
    \item[] Justification: This paper does not involve crowdsourcing or research with human subjects. All experiments are conducted on existing benchmarks using publicly available LLMs.

\item {\bf Institutional review board (IRB) approvals or equivalent for research with human subjects}
    \item[] Question: Does the paper describe potential risks incurred by study participants, whether such risks were disclosed to the subjects, and whether Institutional Review Board (IRB) approvals (or an equivalent approval/review based on the requirements of your country or institution) were obtained?
    \item[] Answer: \answerNA{}
    \item[] Justification: No human subjects are involved in this research.

\item {\bf Declaration of LLM usage}
    \item[] Question: Does the paper describe the usage of LLMs if it is an important, original, or non-standard component of the core methods in this research?
    \item[] Answer: \answerNA{}
    \item[] Justification: LLMs (Llama-3, Qwen3) are the \emph{object of study} in this paper --- we analyze their internal attention statistics --- not a component of the research methodology. LLMs were not used to generate text, make decisions, or produce experimental results.

\end{enumerate}


\newpage
\appendix
\section{Reproducibility Statement}
\label{sec:appendix_reproducibility}

\paragraph{Code and Assets.}
We use publicly available code, models, and datasets, citepd and listed below with corresponding licenses and versions.

\begin{itemize}
\item \textbf{Codebase:} We build upon \texttt{[KVPress]}\url{https://github.com/NVIDIA/kvpress}.
\textit{Version:} 3dbf8f4
\textit{License:} Apache 2.0

\item \textbf{Models:} We use two families of LLMs. 
\begin{enumerate}
    \item \texttt{LLaMA-3} models from~\citep{grattafiori2024llama} of two sizes: 3B, \textit{URL:}~\url{https://huggingface.co/meta-llama/Llama-3.2-3B-Instruct}; and 8B, \textit{URL:}~\url{https://huggingface.co/meta-llama/Llama-3.1-8B-Instruct}. \textit{License:} Llama 3.1 and Llama 3.2 Community License
\item \texttt{Qwen} model of 8B size from \citep{yang2025qwen3technicalreport}. \textit{URL:} \url{https://huggingface.co/Qwen/Qwen3-8B}. \textit{License:} Apache 2.0.
\end{enumerate}

\item \textbf{Datasets:} We use datasets as follows:
\begin{itemize}
\item \textbf{QMSum}~\citep{zhong-etal-2021-qmsum}
\textit{Version:} Latest GitHub release (accessed 2025-05)
\textit{URL:} \url{https://github.com/Yale-LILY/QMSum}
\textit{License:} MIT License

\item \textbf{2WikiMultihopQA}~\citep{ho-etal-2020-constructing}
\textit{Version:} Latest GitHub release (accessed 2025-05)
\textit{URL:} \url{https://github.com/Alab-NII/2wikimultihop}
\textit{License:} Apache 2.0

\item \textbf{SQuAD v2.0}~\citep{rajpurkar2018know}
\textit{Version:} Hugging Face release (accessed 2025-05)
\textit{URL:} \url{https://huggingface.co/datasets/rajpurkar/squad_v2}
\textit{License:} CC BY-SA 4.0

\item \textbf{QuoteSum}~\citep{schuster2023semqa}
\textit{Version:} GitHub release (accessed 2025-05)
\textit{URL:} \url{https://github.com/google-research-datasets/QuoteSum}
\textit{License:} CC BY-SA 4.0

\item \textbf{Verifiability-Granular}~\citep{phukan2024peering}
\textit{Version:} GitHub release (accessed 2025-05)
\textit{URL:} \url{https://github.com/Anirudh-Phukan/verifiability-granular}
\textit{License:} CC BY-SA 4.0
\end{itemize}

\paragraph{Environment.}
Experiments were conducted using PyTorch 2.1, CUDA 12.1 on  A100 80GB, with code and instructions available at \url{[https://anonymous.4open.science/r/RCStat-8E3D/README.md]} for reproducibility.
\end{itemize}


\section{Formalism of Relative Contextualization}
\subsection{Conditional Self and Cross Contextualization}\label{sec:ccc_csc}
The \ref{eq:cc} RV captures the notion of contextualization between the prompt sequence and the generated sequence at a head-level. 
The same for a given sequence $s$ can be formalized as a conditional RV $X|s$, whose Cumulative Distribution Function (CDF) does not include the outermost summation of \ref{eq:cc}. The following definition further qualifies CC to a sub-sequence level.
\begin{definition}[Conditional Cross-Contextualization]\label{def:ccc}
    Assuming a sequence $s=p\oplus g$ with prompt tokens $p$ and generated tokens $g$ is given, and two subsets  $p_1 \subset p$ and $g' \subset g$, whose complementary tokens $p\!\setminus\!p_1$ and $g\!\setminus\!g'$ are given, we define the conditional CC random variable $X_s(p_1, g')$ as
    \begin{align}\label{eq:ccc}
    F_{X_s(p_1\!\leadsto g')}(X_s(p_1\!\leadsto\! g') \leq x) := F_X\!\left(X \leq x \,|\, s, \ p\!\setminus\!p_1,\ g\!\setminus\!g'\right) = \sum_{t_i \in p_1} \sum_{t_j \in g'} \frac{\mathbf{1}(f_s(i,j) \leq x)}{|p_1|\, |g'|}.
\end{align}
\end{definition}
The conditional CC \eqref{eq:ccc} represents the influence of prompt tokens in $p_1$ for generating tokens in $g'$.
Note, if $p_1=\{t_i\}$ and $g'=\{t_j\}$ are singleton sets, then the conditional CC $X_{s}(p_1\!\leadsto \!g')$ is simply a degenerate random variable at $x=f_s(i,j)=\langle q_j, k_i\rangle$. Similarly, if $p_1=p$ and $g'=\{t_j\}$, the conditional CC corresponds to the values in $q_j^TK$, where $K \in \mathbb{R}^{d\times |p|}$ is the matrix of key vectors of prompt tokens. Trivially, $X_s(p\leadsto g) = X|s$.

We define conditional SC similar to conditional CC in Definition \ref{def:ccc}. 
\begin{definition}[Conditional Self-Contextualization]\label{def:csc}
    Assuming a sequence $s=p\oplus g$ is given, and two subsets of the generated tokens $g_1, g' \subset g$, whose complementary tokens $g\!\setminus\!g_1$ and $g\!\setminus\!g'$ are given, we define the conditional SC random variable $Y_s(g_1 \leadsto g')$ as
    \begin{align}\label{eq:csc}
    \!\!F_{Y_s(g_1 \leadsto g')}(Y_s(g_1 \leadsto g') \leq y) := F_Y\!\left(Y\!\leq\!y \,|\, s, g\!\setminus\!g_1, g\!\setminus\!g'\right) = \!\!\!\sum_{t_i\in g_1, t_j\in g'}\! \frac{\mathbf{1}(i \leq j  \wedge f_s(i,j) \leq y)}{\sum_{t_i\in g_1, t_j\in g'} \mathbf{1}(i \leq j)}.
\end{align}
\end{definition}
The conditional SC \eqref{eq:csc} represents the influences of tokens in $g_1$ for generating the tokens in $g'$ that appear after $g_1$. Trivially, $Y_s(g \leadsto g) = Y|s$.

\subsection{Conditional Relative Contextualization}\label{sec:cond_rc}
Due to position embeddings and the auto-regressive nature of generation, self-contextualization values are generally higher than cross-contextualization. Yet some influential prompt tokens also receive higher softmax attention, implying their $\langle q,k\rangle$ values exceed those of certain generated tokens, since softmax is order-preserving. This motivates a third type of variable, \emph{relative contextualization}, designed to isolate prompt-specific influence by subtracting out internal generation bias. Formally, if $X_s(p_1\leadsto g) > Y_s(g_1\leadsto g)$ with high probability, then the (next-layer) value embeddings of the tokens in $g_1$ become heavily affected by the prompt tokens in $p_1$.

\begin{definition}[Conditional Relative Contextualization]\label{def:cond_rc}
    Assuming a sequence $s=p\oplus g$ is given, and two subsets  $p_1\subset p, g_1 \subset g$, whose complementary tokens $p\!\setminus\!p_1$ and $g\!\setminus\!g_1$ are given, we define the relative contextualization (RC) as
    \begin{align}\label{eq:cond_rc}
        Z_s((p_1,g_1)\leadsto g) = \max\left(X_s(p_1\leadsto g) - Y_s(g_1\leadsto g), 0\right).
    \end{align}
\end{definition} Similar to conditional CC \eqref{eq:ccc} and conditional SC \eqref{eq:csc} in \S \ref{sec:ccc_csc}, the conditional random variable $Z_s((p, g) \leadsto g)$ can be written as $Z_s = Z|s = \max (X|s - Y|s, 0)$.
Estimating the central statistics of \ref{eq:rc} requires an operation over the joint distribution of $X$ and $Y$. If there are a large number of prompt and generated tokens, then computing any statistics of $Z$ could become prohibitively expensive, as discussed in \S \ref{sec:comp_complexity}. Note, $X_s$ and $Y_s$ are not independent random variables since the key and query vectors of different tokens are intricately dependent on each other.

In the following, we derive a computationally efficient and practical upper bound for conditional RC \eqref{eq:cond_rc}, without making any distributional assumptions on $X_s$ or $Y_s$. We refer to this upper bound as RC score.

\begin{theorem}[Area Under Conditional CDFs]\label{thm:conditional_oa}
    The expected relative contextualization $Z$, conditioned on two subsequence $p_1$ and $g_1$ as per Definition \ref{def:cond_rc}, is upper bounded by the overlap area $A$ between, a) the area under the marginal CDF $F_Y$ of self-contextualization $Y$, and b) the area under the marginal survival function $S_X$ of cross-contextualization $X$:
    \begin{align}\label{eq:cond_rc_ub}
        \mathbb{E}[Z_s((p_1, g_1)\leadsto g)] \leq A_s((p_1, g_1)\leadsto g) := \int_{-\infty}^{\infty}\!\!\min \left(F_{Y_s(g_1 \leadsto g)}(t),\ S_{X_s(p_1 \leadsto g)}(t)\right) dt, 
    \end{align}
    where $S_X(t)=1-F_X(t)$, and lower bounded by the area $a$, under $F_Y$ but over $F_X$: 
    \begin{align}\label{eq:cond_rc_lb}
    \mathbb{E}[Z_s((p_1,g_1) \leadsto g')] \geq a_s((p_1, g_1) \leadsto g') := \int_{-\infty}^{\infty} \max(F_{Y_s(g_1 \leadsto g)}(t)-F_{X_s(p_1\leadsto g)}, 0) dt. 
    \end{align}
\end{theorem} 

\textbf{Computational Complexity of Conditional RC:} Similar to \eqref{eq:join_rc}, the complexity of computing the expectation of conditional RC is given by
\begin{align}\label{eq:cond_join_rc}
   \mathbb{E}[Z_s((p_1,g_1) \leadsto g')]= \frac{2}{|p_1||g'|^2(|g_1|+1)}\sum_{t_i \in p_1 }\sum_{t_j \in g'} \sum_{t_k \in g_1} \sum_{t_l \in g': l\geq k}\max(f_s(i,j) - f_s(k, l), 0)
\end{align}
\subsection{Theoretical Results}\label{sec:proof}
\begingroup
\renewcommand\thetheorem{\ref{thm:oa}} 
\begin{theorem}[Area Under CDFs]
The expected relative contextualization $Z$ is upper bounded by the overlap area $A$ between, a) the area under the marginal CDF $F_Y$ of self-contextualization $Y$, and b) the area under the marginal survival function $S_X$ of cross-contextualization $X$:
    \begin{align} 
        \mathbb{E}[Z_s(p_1, g_1)] \leq A_s(p_1, g_1) :=  \int_{-\infty}^{\infty}\!\!\min \left(F_{Y_s(g_1, g)}(t),\ S_{X_s(p_1, g)}(t)\right) dt,
    \end{align}
    where $S_X(t)=1-F_X(t)$, and lower bounded by the area $a$, under $F_Y$ but over $F_X$: 
    \begin{align} 
        a_s(p_1, g_1) := \int_{-\infty}^{\infty} \max(F_{Y_s(g_1, g)}(t)-F_{X_s(p_1, g)}, 0) dt 
        \leq \mathbb{E}[Z_s(p_1, g_1)].
    \end{align}
\end{theorem}
\addtocounter{theorem}{-1} 
\endgroup
\begin{proof}
    Following \citep{Vallender1974Wasserstein}, $\kappa = \mathbb{E}[\max(X-Y, 0)]$ 
    can be written as
    \begin{align}
        \kappa &= \int_{-\infty}^\infty P(X > t \text{ and } Y \leq t) dt \\
        &= \int_{-\infty}^\infty 
        \big(P(Y \leq t) - P(X \leq t \text{ and } Y \leq t) \big) dt.
    \end{align}
    Applying Sklar's theorem \citep{Durante2010Copula}, the joint CDF ${P(X \leq t \text{ and } Y \leq t)}$ can be written as a copula $C$ distribution of marginal CDF values:
    \begin{align}
        \kappa = \int_{-\infty}^\infty 
        \big(F_Y(t) - C(F_X(t), F_Y(t)) \big) dt.
    \end{align}
    Let $u\!=\!F_X(t)$ and $v\!=\!F_Y(t)$. Applying Fr\'echet–Hoeffding bound~\citep{Durante2010Copula},
    \begin{alignat}{6}
        &&\max(u + v - 1, 0) &\leq \quad C(u, v) &&\leq \min(u, v) \\
        \Rightarrow && v - \min(u, v) &\leq v -  C(u, v) &&\leq v - \max(u + v - 1, 0) \\
        \Rightarrow && v\!-\!(\min(u\!-\!v,0)\!+\!v) &\leq v - C(u,v) &&\leq v\!-\!(max(u\!-\!1, -v)\!+\!v) \\
        \Rightarrow && \max(v-u,0) &\leq v - C(u,v) &&\leq \min(1-u,v)\label{eq:to_int}.
    \end{alignat}
    We complete the proof by integrating all sides of \eqref{eq:to_int} w.r.t. $t$.
\end{proof}

\begin{corollary}
    For any small $\delta > 0$, the relative contextualization is upper bounded as 
    \begin{align}
        Z_s(p_1,g_1) \leq \frac{A_s(p_1,g_1)}{\delta}
    \end{align}
    with probability at least $1-\delta$.
\end{corollary}
\begin{proof}
    Let \( Z := Z_s(p_1, g_1) \) and \( A := A_s(p_1, g_1) \).  
    From \cref{thm:oa} we know that \( \mathbb{E}[Z] \leq A \).

    Now, by Markov's inequality, for any \( \delta > 0 \), we have:
    \begin{align*}
        \Pr\left(Z \geq \frac{A}{\delta} \right) \leq \frac{\mathbb{E}[Z]}{A/\delta} \leq \frac{A}{A/\delta} = \delta.
    \end{align*}
    Therefore, with probability at least \( 1 - \delta \), we have:
    \begin{align*}
        Z < \frac{A}{\delta}.
    \end{align*}
    Hence,
    \begin{align*}
        Z_s(p_1, g_1) \leq \frac{A_s(p_1, g_1)}{\delta}
    \end{align*}
    with probability at least \( 1 - \delta \), as claimed.
\end{proof}

\newpage
\section{Analysis of RC Scores of Attention heads}
\subsection{Two contrasting examples}
\label{sec:appendix_prompt_examples}
\begin{figure}[t]
  \centering
\includegraphics[width=0.55\linewidth]{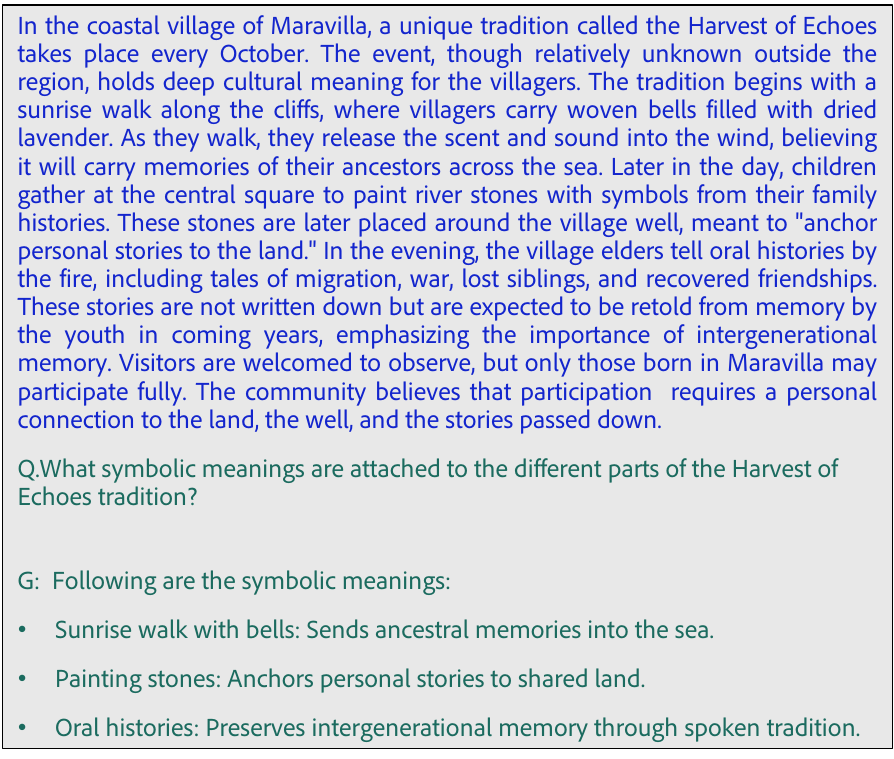}
  \caption{Example 1: Relevant context}
  \label{fig: Prompt 1}
\end{figure}

\begin{figure}[H]
  \centering
\includegraphics[width=0.85\linewidth]{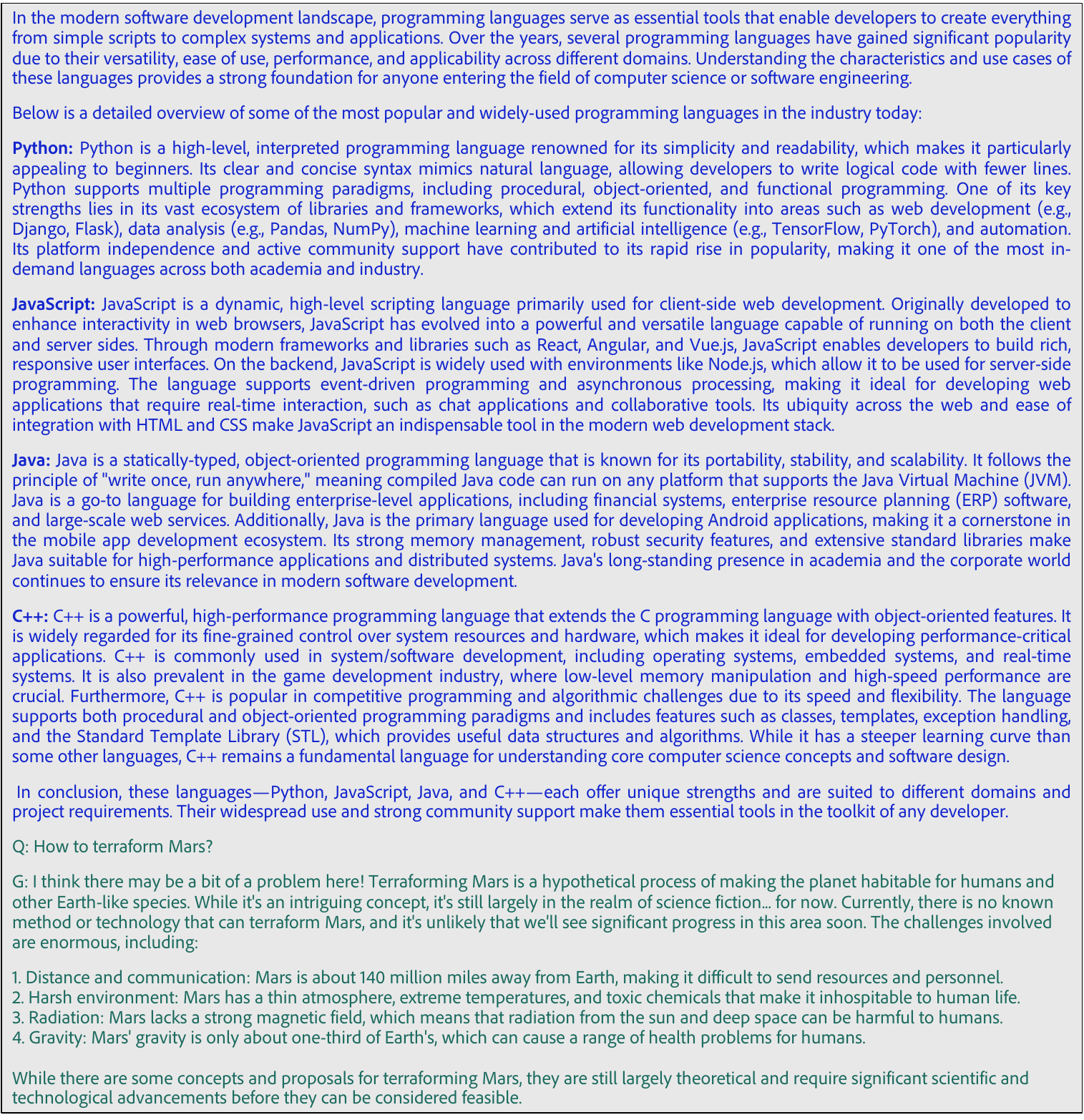}
  \caption{Example 2: Irrelevant context}
  \label{fig: Prompt 2}
\end{figure}

\newpage

\begin{figure}[H]
\centering
\begin{subfigure}{0.3\textwidth}
    \centering
    \includegraphics[width=\linewidth]{Figures/qmsum_erc_consistency.pdf}
    \caption{QMSum dataset}
    \label{fig:qmsum_erc_app}
\end{subfigure}~
\begin{subfigure}{0.3\textwidth}
    \centering
    \includegraphics[width=\linewidth]{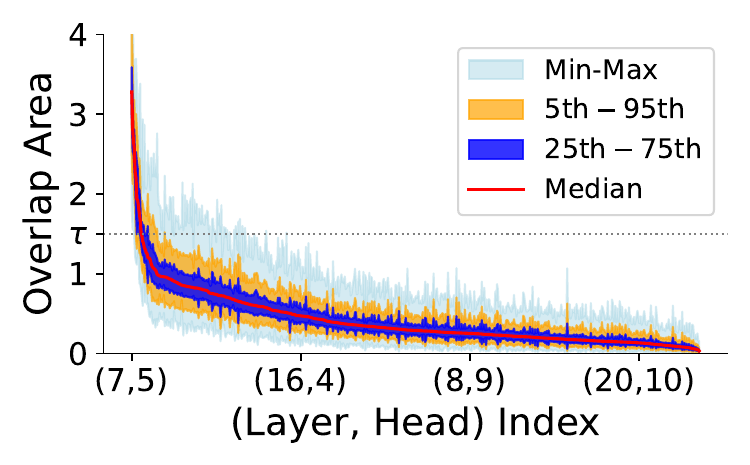}
    \caption{2WikiMultihopQA dataset}
    \label{fig:2WikiMultihopQA_erc_app}
\end{subfigure}~
\begin{subfigure}{0.3\textwidth}
    \centering
    \includegraphics[width=\linewidth]{Figures/squad_erc_consistency.pdf}
    \caption{SQuAD dataset}
    \label{fig:squad_erc_app}
\end{subfigure}
\caption{\label{fig:head_roles_app} All plots are generated using the mentioned datasets with LLaMA-3B model: 28 layers, 24 heads in each layer. All plots show the RC score of the heads (upper bound of expected RC).
}
\vspace{-5mm}
\end{figure}~
\begin{figure}[H]
\centering
\begin{subfigure}{0.3\textwidth}
    \centering
    \includegraphics[width=\linewidth]{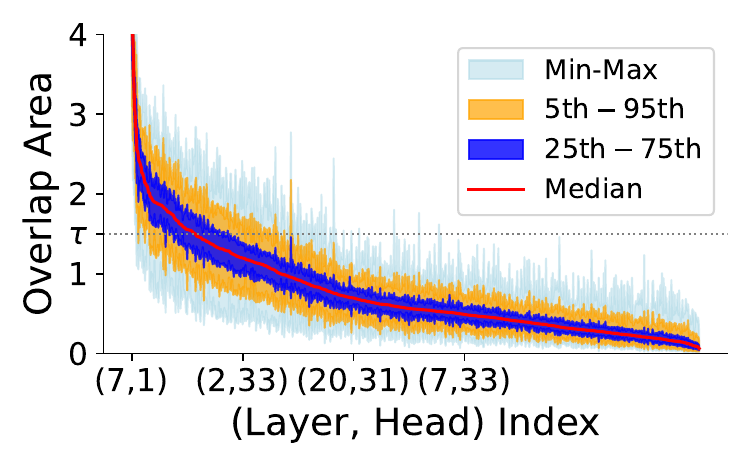}
    \caption{QMSum dataset}
\end{subfigure}~
\begin{subfigure}{0.3\textwidth}
    \centering
    \includegraphics[width=\linewidth]{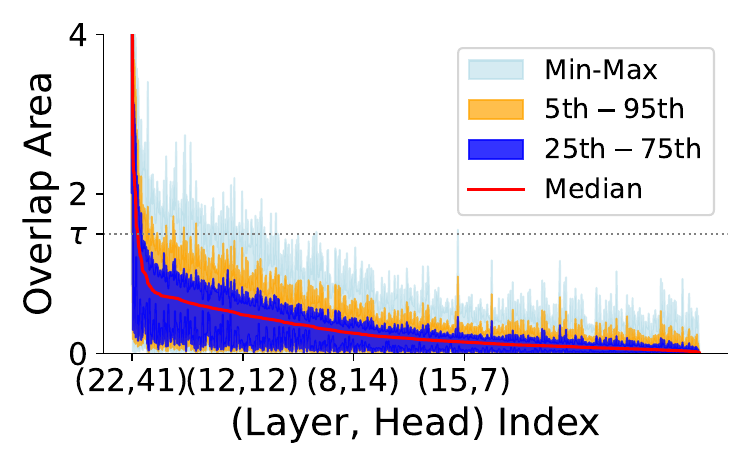}
    \caption{2WikiMultihopQA dataset}
\end{subfigure}~
\begin{subfigure}{0.3\textwidth}
    \centering
    \includegraphics[width=\linewidth]{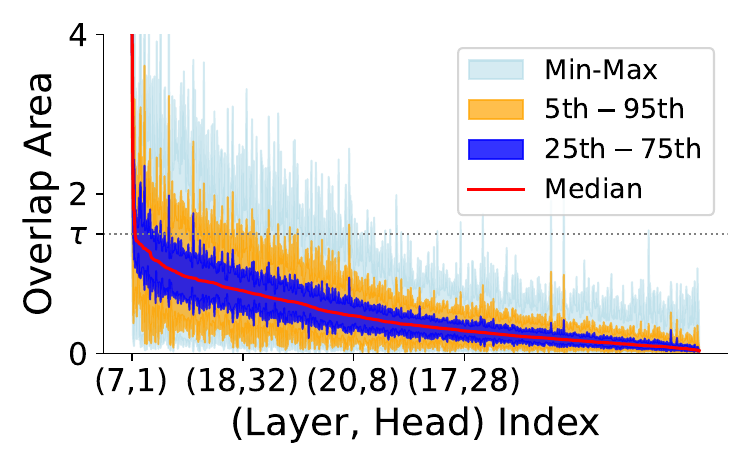}
    \caption{SQuAD dataset}
\end{subfigure}
\caption{\label{fig:head_roles_llama8b_app} All plots are generated using the mentioned datasets with LLaMA-8B model: 32 layers, 32 heads in each layer. All plots show the RC score of the heads (upper bound of expected RC).
}
\vspace{-5mm}
\end{figure}~
\begin{figure}[H]
\centering
\begin{subfigure}{0.3\textwidth}
    \centering
    \includegraphics[width=\linewidth]{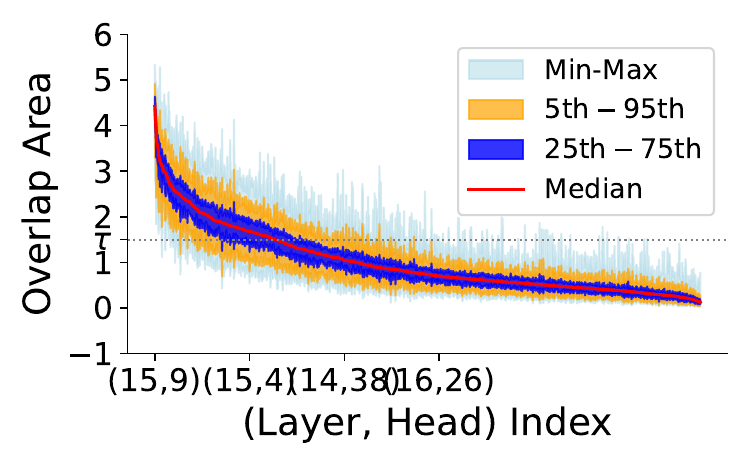}
    \caption{QMSum dataset}
\end{subfigure}~
\begin{subfigure}{0.3\textwidth}
    \centering
    \includegraphics[width=\linewidth]{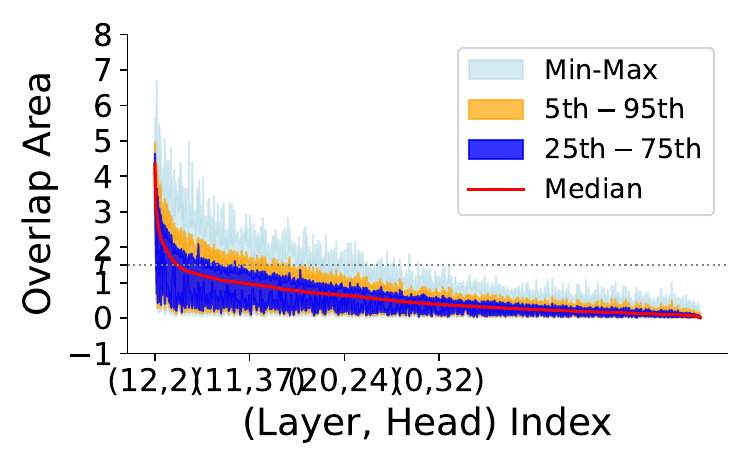}
    \caption{2WikiMultihopQA dataset}
\end{subfigure}~
\begin{subfigure}{0.3\textwidth}
    \centering
    \includegraphics[width=\linewidth]{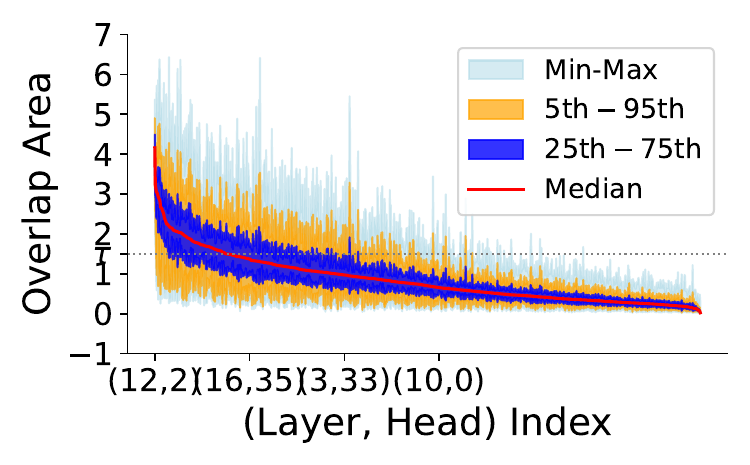}
    \caption{SQuAD dataset}
\end{subfigure}
\caption{\label{fig:head_roles_qwen_app} All plots are generated using the mentioned datasets with Qwen3-8B model: 36 layers, 32 heads in each layer. All plots show the RC score of the heads (upper bound of expected RC).
}
\vspace{-5mm}
\end{figure}
\subsection{Sensitivity of RCStat to Task Complexity}\label{sec:task_complexity}
To better understand the behavior of RCStat across tasks of varying complexity, we compare the RC scores of the heads in LLaMA-3B model in \cref{fig:head_roles_app} on single-hop question answering (SQuAD v2), multi-hop question answering (2WikiMultiHopQA), and multi-sentence summarization (QMSum). We apply the same $\tau$-threshold analysis ($\tau =$ 1.5) to quantify the number of attention heads with RC $\geq \tau$.

\textbf{Single-hop QA (SQuAD v2)}: The median number of heads $\geq \tau$ is 7. For example, in the question “Murders were the base for which story that Capote wrote?”, the answer resides in a single span. Only a small number of heads require high RC, reflecting localized retrieval.

\textbf{Multi-hop QA (2WikiMultiHopQA):} The median rises to 15, approximately 2$\times$ that of single-hop QA. For instance, in “Where was the place of death of the director of the film Happy Hobos?”, the model must first identify the director’s name and then locate their place of death—two distinct spans that must be aggregated. This leads to a greater number of heads with high RC.

\textbf{Summarization (QMSum):} The median further increases to 62, as summarization requires integrating information across the entire document. A large number of heads exhibit high RC, consistent with the need for broad contextualization.

This progression (7 → 15 → 62) highlights that RCStat naturally scales with the breadth of contextualization, ranging from localized lookup (few heads in single-hop), to multi-span aggregation (moderate number of heads in multi-hop), to full-document comprehension (many heads in summarisation).

Importantly, this also demonstrates how RCStat does not require predefining which or how many heads are relevant. Instead, it adapts automatically to the input task, calibrating head selection based on contextual requirements. While RC does not directly measure logical reasoning, it provides a faithful proxy for quantifying the extent of contextual integration needed across tasks of increasing complexity.

\subsubsection{Persistence of top RC-informative heads across models
} 
\cref{fig:head_roles_llama8b_app} \cref{fig:head_roles_qwen_app} show similar power law pattern of RC scores for LLaMA-8B and Qwen-8B models across all the datasets, indicating that this a generalizable phenomenon and artifact of model pre-training. 

\subsection{Relative Contextualization distribution: Head-wise Analysis} \label{sec:RC_distribution}
We show the per-head distribution of relative contextualization, in terms of the upper bound overlap area, in Figures \ref{fig:rc_dist_qmsum}, \ref{fig:rc_dis_squad}, and \ref{fig:rc_dis_2wiki} for QMSum, Squadv2, and 2WikiMultiHop datasets, respectively. The percentile values of these distributions are shown in Figures \ref{fig:percentile_rc_qmsum}, \ref{fig:percentile_rc_squad}, and \ref{fig:percentile_rc_2wiki} respectively. These figures provide empirical evidence of our statement in the conclusion section: 
``the most influential contextualization heads consistently reside in the model’s middle layers, corroborating prior findings.'' This can also be observed in Figures 4c and 4d in the main paper, where the high-scoring heads correspond to the middle layers: layer indices are shown in the x-axis labels. 




\section{Experiments for KV Compression}
\subsection{Evaluation Under Varied Decoding Conditions}\label{sec:decoding_strategy}

Robustness to different decoding strategies (e.g., greedy decoding, top-k sampling with temperature, or beam search) is an important consideration for evaluating KV-compression. Conventional textual metrics such as BLEU or ROUGE can be sensitive to these decoding variations. 

To address this, our primary metric is VER (Value Error Rate), which directly compares the final-layer hidden states of the response/generated tokens when they are obtained with and without KV eviction. It does not perform auto-regressive decoding; instead, it executes a multi-token forward pass of the ground-truth response tokens, with the KV caches of the prompt tokens evicted. Here, the error cascades across the layers during the forward pass, and the final layer hidden states are compared against the final layer hidden states when there is no KV eviction.
This approach decouples compression quality from decoding artifacts, ensuring more reliable comparisons across decoding settings. To capture the error cascading across the auto-regressive decoding, we resort to conventional textual metrics, such as ROUGE-1 and ROUGE-L, using the simple greedy decoding strategy. 

\textbf{Note:} For the attribution usecase in \S \ref{sec:exp_attribution}, decoding variations are not relevant. Attribution is performed post-generation on the fixed prompt tokens and response tokens as provided by different benchmark datasets.

\subsection{KV Compression results on Llama Models}
\label{sec:llama_results_kvc}

This appendix presents all evaluation plots across datasets (QMSum, 2WikiMultiHop, SQuAD) and model sizes (3B, 8B). Each plot compares performance metrics--VER, ROUGE-1 F1, and ROUGE-L F1-- across different configurations. A shared legend is included for clarity, and plots are grouped by metric and model size for visual consistency.

\begin{figure}[H]
    \centering
    \begin{subfigure}{\textwidth}
    \centering
        \includegraphics[width=0.8\linewidth]{Figures/only_legend.pdf}
    \end{subfigure}\\[-2mm]
    \begin{subfigure}{0.33\textwidth}
        \includegraphics[width=\linewidth]{Figures/dv_qmsum_3B.pdf}
        \label{fig:appx_dv_qmsum}
    \end{subfigure}~
    \begin{subfigure}{0.33\textwidth}
        \includegraphics[width=\linewidth]{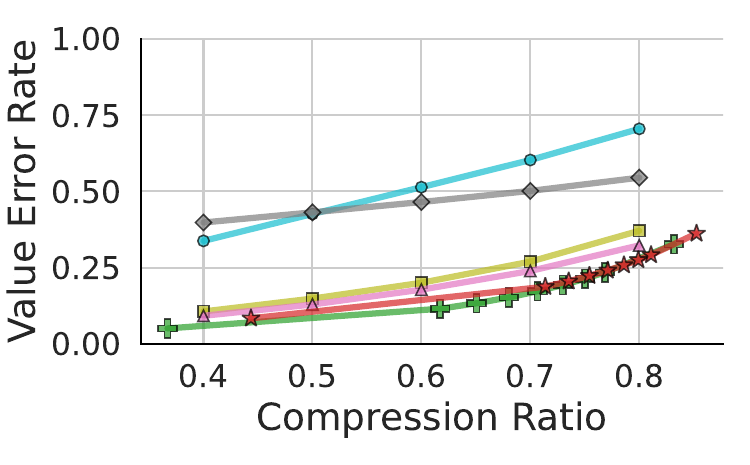}
        \label{fig:appx_r1f1_qmsum}
    \end{subfigure}~
    \begin{subfigure}{0.33\textwidth}
        \includegraphics[width=\linewidth]{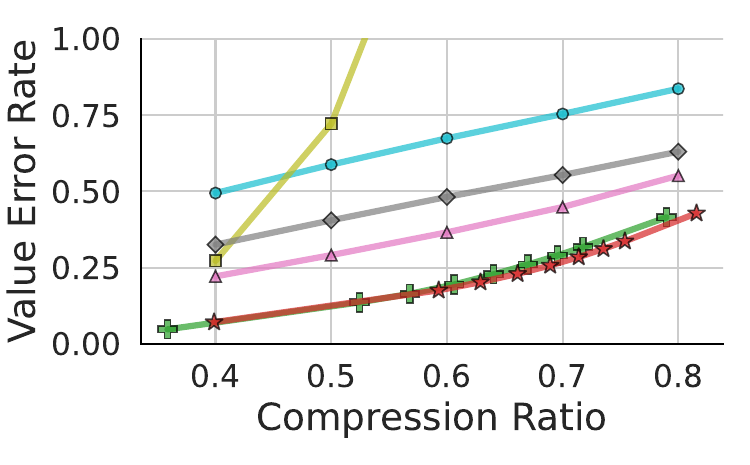}
        \label{fig:appx_rlf1_qmsum}
    \end{subfigure}
    \caption{\label{fig:appx_kv_comp_qmsum} VER scores for 3B model on QMSum, 2WikiMultiHop, and SQuAD datasets.}
\end{figure}

\begin{figure}[H]
    \centering
    \begin{subfigure}{\textwidth}
    \centering
        \includegraphics[width=0.8\linewidth]{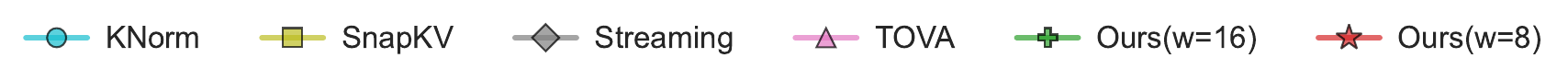}
    \end{subfigure}\\[-2mm]
    \begin{subfigure}{0.33\textwidth}
        \includegraphics[width=\linewidth]{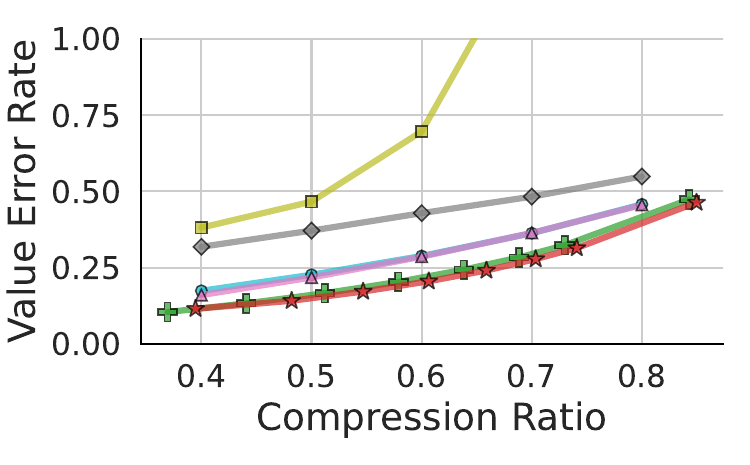}
        \label{fig:appx_dv_qmsum_8B}
    \end{subfigure}~
    \begin{subfigure}{0.33\textwidth}
        \includegraphics[width=\linewidth]{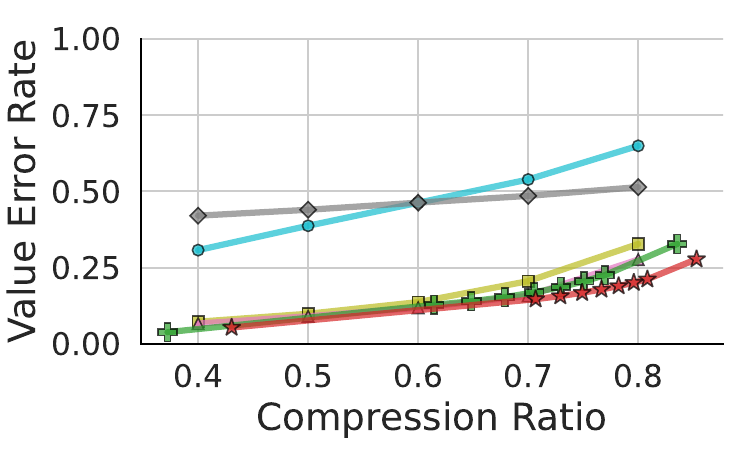}
        \label{fig:appx_dv_2wiki_8B}
    \end{subfigure}~
    \begin{subfigure}{0.33\textwidth}
        \includegraphics[width=\linewidth]{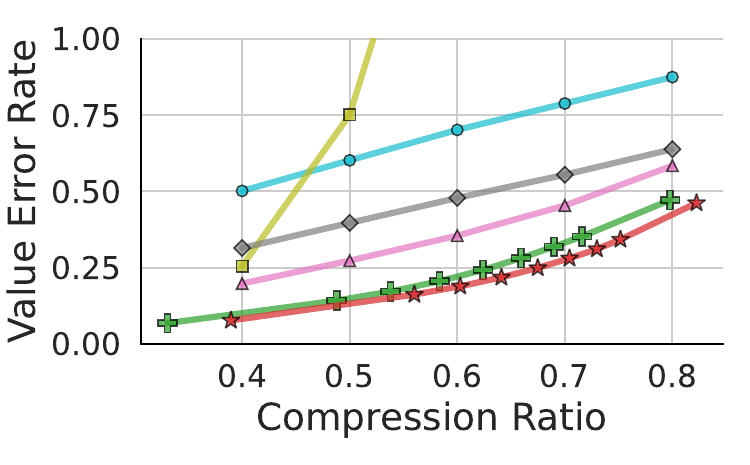}
        \label{fig:appx_dv_squad_8B}
    \end{subfigure}
    \caption{\label{fig:appx_kv_comp_dv} VER scores for 8B model on QMSum, 2WikiMultiHop, and SQuAD datasets.}
\end{figure}

\begin{figure}[H]
    \centering
    \begin{subfigure}{\textwidth}
    \centering
        \includegraphics[width=0.8\linewidth]{Figures/only_legend.pdf}
    \end{subfigure}\\[-2mm]
    \begin{subfigure}{0.33\textwidth}
        \includegraphics[width=\linewidth]{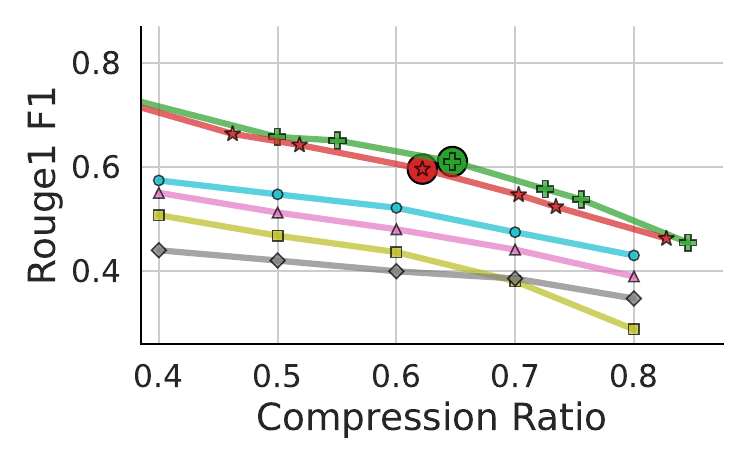}
        \label{fig:appx_r1f1_qumsum_3B}
    \end{subfigure}~
    \begin{subfigure}{0.33\textwidth}
        \includegraphics[width=\linewidth]{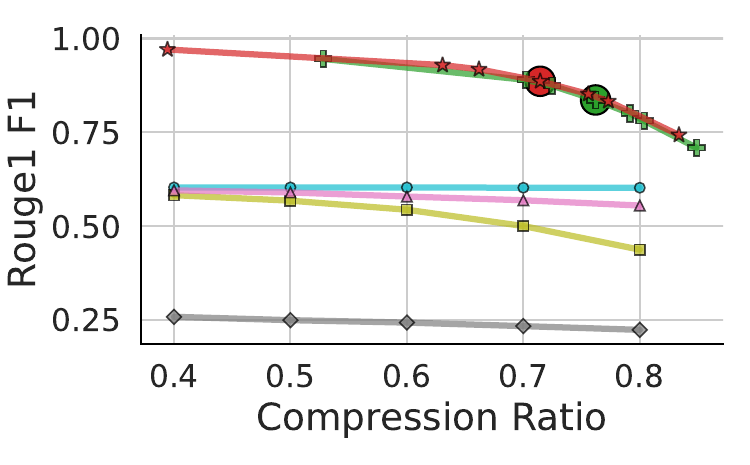}
        \label{fig:appx_r1f1_2wiki_3B}
    \end{subfigure}~
    \begin{subfigure}{0.33\textwidth}
        \includegraphics[width=\linewidth]{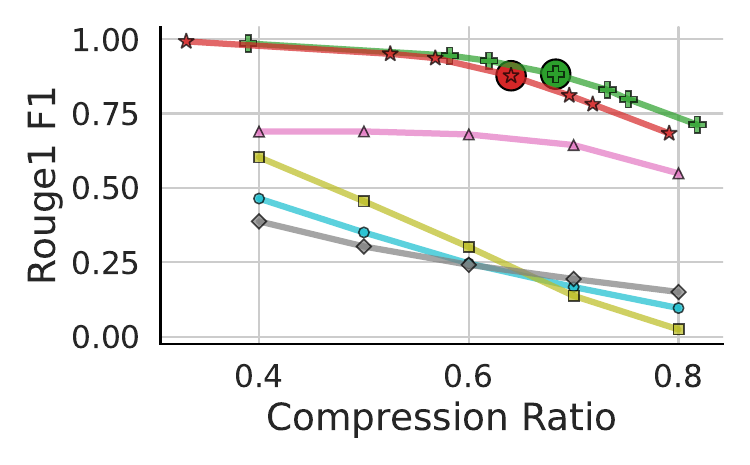}
        \label{fig:appx_r1f1_squad_3B}
    \end{subfigure}
    \caption{\label{fig:appx_kv_comp_r1_3b} ROUGE-1 F1 scores for 3B model on QMSum, 2WikiMultiHop, and SQuAD datasets.}
\end{figure}

\begin{figure}[H]
    \centering
    \begin{subfigure}{\textwidth}
    \centering
        \includegraphics[width=0.8\linewidth]{Figures/only_legend.pdf}
    \end{subfigure}\\[-2mm]
    \begin{subfigure}{0.33\textwidth}
        \includegraphics[width=\linewidth]{Figures/rLf1_qmsum_3B.pdf}
        \label{fig:appx_rLf1_qumsum_3B}
    \end{subfigure}~
    \begin{subfigure}{0.33\textwidth}
        \includegraphics[width=\linewidth]{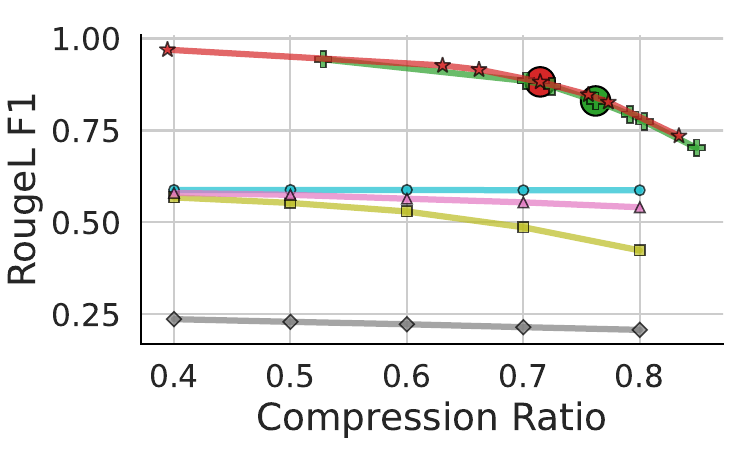}
        \label{fig:appx_rLf1_2wiki_3B}
    \end{subfigure}~
    \begin{subfigure}{0.33\textwidth}
        \includegraphics[width=\linewidth]{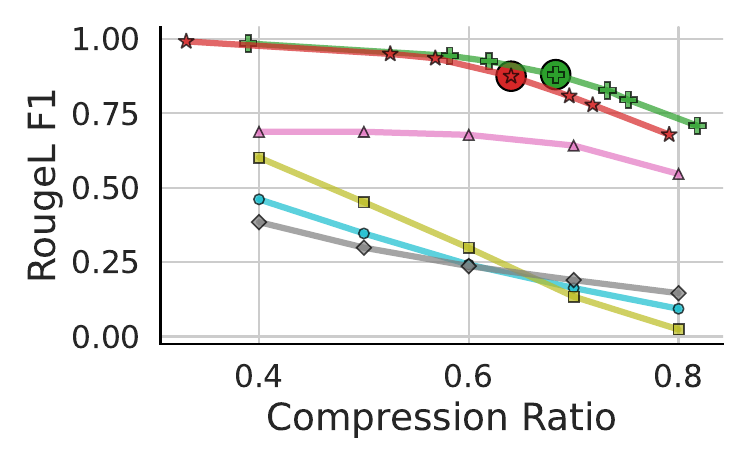}
        \label{fig:appx_rLf1_squad_3B}
    \end{subfigure}
    \caption{\label{fig:appx_kv_comp_rl_3b} ROUGE-L F1 scores for 3B model on QMSum, 2WikiMultiHop, and SQuAD datasets.}
\end{figure}

\begin{figure}[H]
    \centering
    \begin{subfigure}{\textwidth}
    \centering
        \includegraphics[width=0.8\linewidth]{Figures/only_legend.pdf}
    \end{subfigure}\\[-2mm]
    \begin{subfigure}{0.33\textwidth}
        \includegraphics[width=\linewidth]{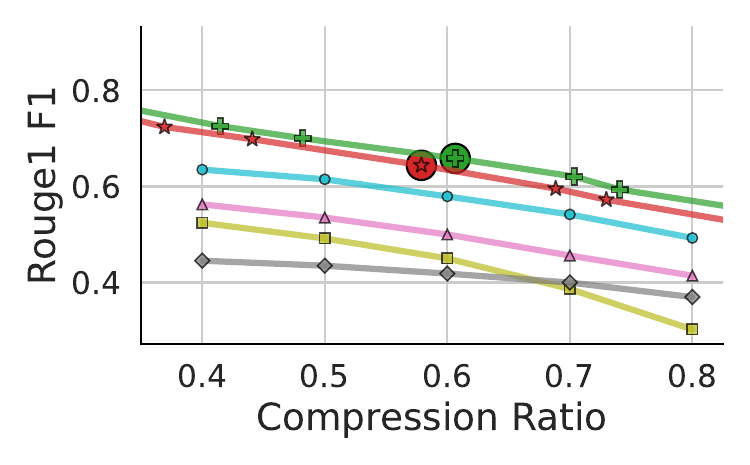}
        \label{fig:appx_r1f1_qmsum_8B}
    \end{subfigure}~
    \begin{subfigure}{0.33\textwidth}
        \includegraphics[width=\linewidth]{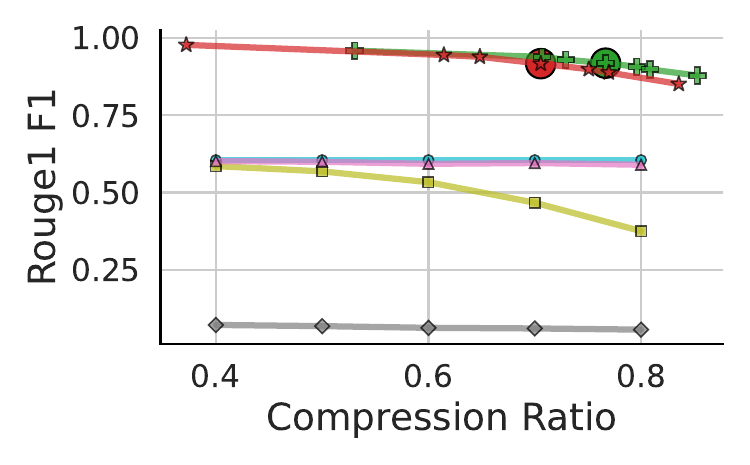}
        \label{fig:appx_r1f1_2wiki_8B}
    \end{subfigure}~
    \begin{subfigure}{0.33\textwidth}
        \includegraphics[width=\linewidth]{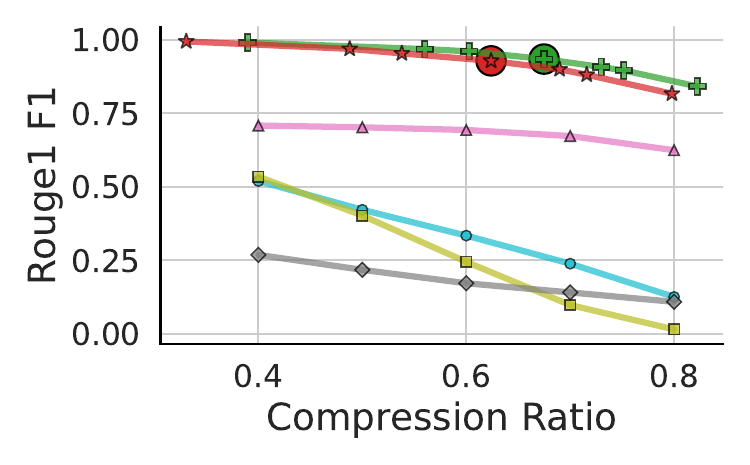}
        \label{fig:appx_r1f1_squad_8B}
    \end{subfigure}
    \caption{\label{fig:appx_kv_comp_r1_8b} ROUGE-1 F1 scores for 8B model on QMSum, 2WikiMultiHop, and SQuAD datasets.}
\end{figure}

\begin{figure}[H]
    \centering
    \begin{subfigure}{\textwidth}
    \centering
        \includegraphics[width=0.8\linewidth]{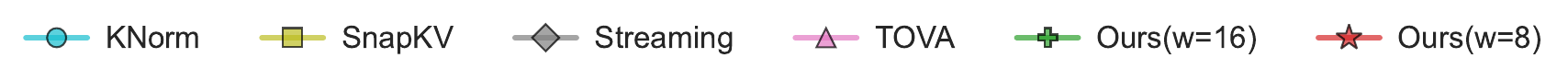}
    \end{subfigure}\\[-2mm]
    \begin{subfigure}{0.33\textwidth}
        \includegraphics[width=\linewidth]{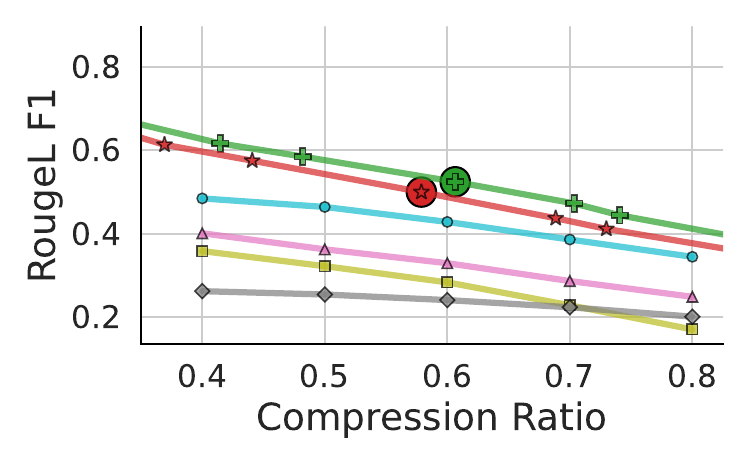} 
        \label{fig:appx_rLf1_qmsum_8B}
    \end{subfigure}~
    \begin{subfigure}{0.33\textwidth}
        \includegraphics[width=\linewidth]{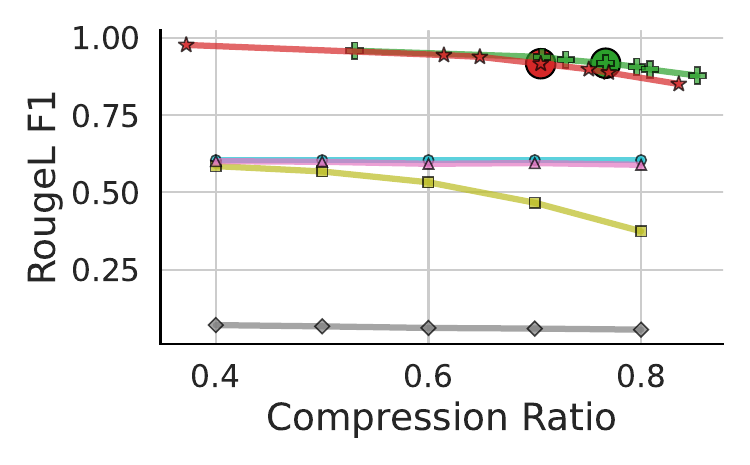}
        \label{fig:appx_rLf1_2wiki_8B}
    \end{subfigure}~
    \begin{subfigure}{0.33\textwidth}
        \includegraphics[width=\linewidth]{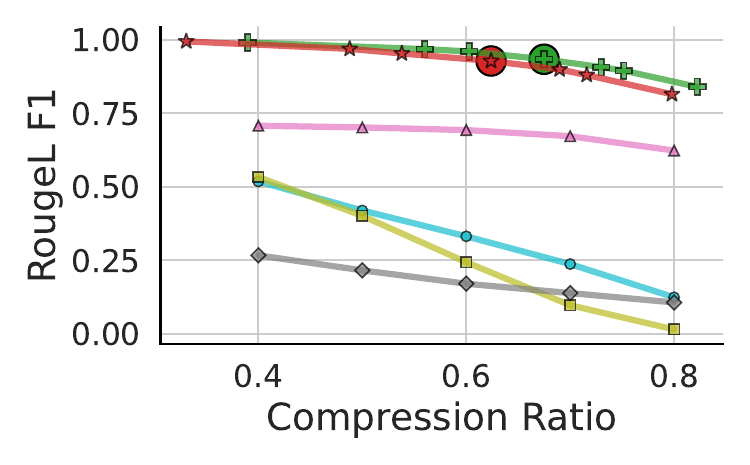}
        \label{fig:appx_rLf1_squad_8B}
    \end{subfigure}
    \caption{\label{fig:appx_kv_comp_rl_8b} ROUGE-L F1 scores for 8B model on QMSum, 2WikiMultiHop, and SQuAD datasets.}
\end{figure}

\begin{table}[h]
\centering
\caption{
\textbf{Value Error Rate (VER)} on the QMSum dataset across different compression ratios (50\%, 60\%, 70\%) for LLaMA-3.2-3B and LLaMA-3.1-8B Instruct models. \textbf{RCStat (IOT)} assumes \textit{Independent Output Tokens}, while \textbf{RCStat (Non-IOT)} does not assume any independence. Here, lower is better.
}\label{tab:ver-qmsum}
\vspace{0.5em}
\begin{tabular}{llccc}
\toprule
\textbf{Model} & \textbf{Method} & \textbf{50\%} & \textbf{60\%} & \textbf{70\%} \\
\midrule
\multirow{3}{*}{\textbf{3B}} 
& TOVA               & 0.2408 & 0.3103 & 0.3905 \\
& \textsc{RCStat} (IOT)       & 0.1956 & 0.2648 & 0.3402 \\
\rowcolor{mygreen} & \textsc{RCStat} (Non-IID)   & \textbf{0.1571} & \textbf{0.2295} & \textbf{0.3066} \\
\midrule
\multirow{3}{*}{\textbf{8B}} 
& TOVA               & 0.2177 & 0.2859 & 0.3639 \\
& \textsc{RCStat} (IOT)       & 0.1615 & 0.2290 & 0.3007 \\         
\rowcolor{mygreen} & \textsc{RCStat} (Non-IID)   & \textbf{0.1043} & \textbf{0.2034} & \textbf{0.2836} \\
\bottomrule
\end{tabular}
\end{table}

\subsection{Generalizability of RCStat Across Model Families}\label{sec:qwen_results_kvc}
To demonstrate that RCStat is not specific to the LLaMA series, we repeated core evaluations on the \textbf{Qwen3-8B} model. Results on the QMSum summarization task confirm that Relative Contextualization properties hold across different model families. RCStat achieves up to \textbf{36\% lower VER} than TOVA and over \textbf{70\% lower} than KNorm/SnapKV, validating strong generalization. Similar trends are observed on SQuAD and 2Wiki, where RCStat consistently outperforms all baselines across compression ratios, further establishing its robustness beyond a single dataset.

\begin{table}[h]
\centering
\scriptsize
\caption{Value Error Rate (VER) with Qwen3-8B for different KV compression methods across three datasets. RCStat consistently shows strong performance across Compression Ratios (CR).}
\begin{tabular}{l|ccccc|ccccc|ccccc}
\toprule
& \multicolumn{5}{c|}{\textbf{QMSum}} & \multicolumn{5}{c|}{\textbf{2Wiki}} & \multicolumn{5}{c}{\textbf{SQuAD v2}} \\
\cmidrule(lr){2-6} \cmidrule(lr){7-11} \cmidrule(lr){12-16}
\textbf{Method \hfill CR} & 0.40 & 0.50 & 0.60 & 0.70 & 0.80 & 
                 0.40 & 0.50 & 0.60 & 0.70 & 0.80 & 
                 0.40 & 0.50 & 0.60 & 0.70 & 0.80 \\
\midrule
RCStat ($w=16$) & \textbf{0.12} & \textbf{0.16} & \textbf{0.22} & \textbf{0.31} & \textbf{0.47} &
                  \textbf{0.13} & \textbf{0.18} & \textbf{0.24} & \textbf{0.32} & \textbf{0.46} &
                  \textbf{0.09} & \textbf{0.13} & \textbf{0.18} & \textbf{0.26} & \textbf{0.39} \\
RCStat ($w=8$)  & \textbf{0.13} & \textbf{0.16} & \textbf{0.21} & \textbf{0.29} & \textbf{0.45} &
                  \textbf{0.14} & \textbf{0.19} & \textbf{0.25} & \textbf{0.33} & \textbf{0.48} &
                  \textbf{0.10} & \textbf{0.14} & \textbf{0.19} & \textbf{0.27} & \textbf{0.41} \\
TOVA             & 0.19 & 0.21 & 0.25 & 0.33 & 0.49 &
                  0.20 & 0.24 & 0.29 & 0.37 & 0.53 &
                  0.16 & 0.19 & 0.23 & 0.30 & 0.46 \\
KNorm            & 0.39 & 0.41 & 0.48 & 0.61 & 0.80 &
                  0.40 & 0.45 & 0.52 & 0.65 & 0.85 &
                  0.32 & 0.37 & 0.44 & 0.58 & 0.77 \\
SnapKV           & 0.45 & 0.55 & 0.72 & 1.34 & 2.33 &
                  0.47 & 0.58 & 0.75 & 1.30 & 2.25 &
                  0.41 & 0.52 & 0.69 & 1.20 & 2.10 \\
Streaming        & 0.26 & 0.31 & 0.37 & 0.43 & 0.54 &
                  0.25 & 0.30 & 0.36 & 0.44 & 0.56 &
                  0.21 & 0.27 & 0.34 & 0.41 & 0.51 \\
\bottomrule
\end{tabular}
\end{table}

\noindent
\textbf{Note:} RCStat uses an adaptive compression ratio, but interpolated VER scores at fixed CRs are reported here for comparability with the Figures presented in the main paper.

\subsection{KV-Compression Efficiency Analysis}\label{sec:efficiency_kvc}

In addition to accuracy under fixed compression ratios, efficiency is a critical factor in evaluating KV-compression methods. Efficiency can be viewed along two complementary axes: \textbf{memory savings} and \textbf{computational latency}.

\textbf{Memory savings.} KV memory footprint is often the dominant bottleneck in real-world deployments. Compression that maintains answer quality while reducing memory is, therefore, a central challenge. RCStat achieves significant improvements in this dimension, providing up to 32\% higher memory savings than the best-performing baseline at equivalent accuracy levels (Figure 6c).

\textbf{Computational latency.} Latency for compression-based decoding can be decomposed into:

\begin{enumerate}
    \item \textbf{Prefill:} This step remains unchanged across all methods and dominates runtime (~3 ms per layer on Llama-3.2-3B for QMSum with $\approx$1000 average prefill length).
    \item \textbf{Compression decision:} RCStat requires 0.68 ms per layer, which is on par with SNAP (0.66 ms) and TOVA (0.65 ms), while remaining only modestly slower than KNorm (0.13 ms) and STREAM (0.15 ms). Since this step can be pipelined with layer-wise prefill computation, the effective overhead is negligible.
    \item \textbf{Decode:} Latency improvements during decoding scale directly with compression ratio. Because RCStat maintains higher answer quality even under more aggressive compression, it enables proportionally greater decoding speed-ups.
\end{enumerate}

Taken together, these results show that RCStat offers the strongest balance of memory efficiency and computational speed, outperforming prior methods on memory savings while achieving comparable per-layer latency to state-of-the-art baselines. We will release a head-wise compression implementation to support efficient decoding with RCStat in future work.

\subsection{Effect of Window Size on KV Compression}\label{sec:window_size}

In KV cache compression, the ``window'' refers to the last few prompt tokens used as a proxy for future tokens during inference. It is important to clarify that this is not a sliding window. We evaluated window sizes $w \in \{1, 2, 4, 8, 16, 32, 64\}$ on the Llama3.1-8B model across different compression ratios (CR), measuring Value Error Rate (VER) as in \cref{fig:dv_qmsum_8B}. For each window size, the compression factor in Equation \eqref{eq:eviction} was varied from 0.02 to 2 to obtain different CRs.  

We observe from the results in \cref{tab:ws_variation} that window sizes $w=8$ and $w=16$ consistently lie on the Pareto frontier, achieving a balance between compression and accuracy. Smaller windows ($w=1,2$) lead to higher VER despite higher compression, while larger windows ($w \geq 32$) yield lower compression ratios. This suggests a ``sweet spot'' in window size that balances context coverage and relevance.

\begin{table}[h]
\centering
\scriptsize
\caption{VER under varying window sizes ($w$) and compression ratios (CR) for KV cache compression. Bold numbers indicate Pareto-optimal results.}\label{tab:ws_variation}
\resizebox{\columnwidth}{!}{
\begin{tabular}{c|cccccc|cccccc|cccccc}
\toprule
 & \multicolumn{6}{c|}{QMSum} &  \multicolumn{6}{c|}{SQuAD v2} &  \multicolumn{6}{c}{2WikiMultiHopQA} \\
\cmidrule(lr){2-7} \cmidrule(lr){8-13} \cmidrule(lr){14-19}
$w$ & 0.4 & 0.5 & 0.6 & 0.7 & 0.8 & 0.9 &  0.4 & 0.5 & 0.6 & 0.7 & 0.8 & 0.9 &  0.4 & 0.5 & 0.6 & 0.7 & 0.8 & 0.9 \\
\midrule
1 & - & - & - & - & - & 0.74 &  - & - & - & - & - & 1.00 &  - & - & - & - & - & 0.92 \\
2 & - & - & - & - & 0.60 & 0.63 &  - & - & - & - & 0.51 & 0.63 &  - & - & - & - & - & 0.48 \\
4 & - & - & 0.27 & 0.36 & 0.42 & 0.52 &  - & - & 0.25 & 0.29 & 0.44 & 0.60 &  - & - & 0.14 & 0.19 & 0.24 & 0.36 \\
8 & 0.12 & \textbf{0.15} & \textbf{0.20} &\textbf{ 0.28} & \textbf{0.40} & \textbf{0.51} &  - & \textbf{0.13} & \textbf{0.19} & \textbf{0.28} & \textbf{0.42} & \textbf{0.59} & {0.05} & \textbf{0.07} & \textbf{0.11} & \textbf{0.14} & \textbf{0.20} & \textbf{0.35} \\
16 & \textbf{0.11} & 0.16 & 0.22 & \textbf{0.28} & 0.42 & 0.54 & \textbf{0.10} & 0.15 & 0.22 & 0.33 & 0.45 & 0.61 &  \textbf{0.04} & 0.08 & 0.12 & 0.17 & 0.27 & 0.42 \\
32 & 0.14 & 0.19 & 0.26 & 0.34 & 0.43 & -  & 0.21 & 0.29 & 0.38 & 0.47 & - & - &  0.13 & 0.18 & 0.24 & 0.31 & 0.38 & - \\
64 & 0.13 & 0.19 & 0.26 & 0.35 & - & - &  0.22 & 0.31 & 0.40 & - & - & - &  0.14 & 0.19 & 0.25 & 0.32 & - & - \\
\bottomrule
\end{tabular}}
\end{table}

\subsection{Independence assumption of generated tokens}\label{sec:independence}
An even simpler approximation is to use conditional expectation by assuming output tokens are independent of each other and uniformly distributed: 
\begin{align}\label{eq:iid}
    \mathbb{E}[Z|s] \approx \mathbb{E}_{t_{i}\sim g}[\mathbb{E}[Z|s, \{t_i\}] = \mathbb{E}_{t_{i}\sim g}[\mathbb{E}[\max(X_s(p\leadsto \{t_j\}) - Y_s(g\leadsto \{t_j\}, 0)]].
\end{align}

The result in \cref{tab:ver-qmsum} shows that the fidelity of value vectors is higher when \textsc{RCStat} is executed without assuming independence for the random variables corresponding to $\langle q, k \rangle$ of generated tokens. Nonetheless, even with the independence assumption, \textsc{RCStat} outperforms TOVA, which is the best-performing method in our experiments for the main paper.


\subsection{Complete Experimental Statistics for KV-compression results} \label{sec:supp_material}
Please find the results of Value Error Rate (VER) inside the \texttt{VER} folder. For the baseline methods, the mean and standard deviations of VER for different compression ratios are saved in \texttt{csv} files with the naming format \texttt{<dataset>\_<model>\_baseline\_df.csv}, where the dataset field can be \texttt{2WikiMultiHop}, \texttt{QMSum}, or \texttt{SQuAD}, and the model field can be \texttt{3b} or \texttt{8b}. 
Similarly, the mean and standard deviations of VER and the mean and standard deviations of the compression ratios for different threshold multipliers are saved in \texttt{csv} files with the naming format \texttt{<dataset>\_<model>\_proposed\_df.csv}. Similarly, the results for Rouge1 and RougeL can be found in the \texttt{All\_Rouges} folder.

\section{Additional Results for Post-hoc Attribution}

\subsection{Post vs Pre soft-max RC}\label{sec:post_vs_pre}

Using LLaMa-3.1 8B model we demonstrate how selecting heads based on post-softmax RC scores performs worse than all heads as shown in \cref{tab:table_attribution_pre_post} in \S \ref{sec:post_vs_pre}.
This observation is also generalizable to all other LLMs as well as with different sizes. 
This shows the importance performing RC analysis using pre-softmax as otherwise it can lead to sub-optimal head selection. 

\begin{table}[h]
    \caption{Chunk-level accuracy for attributing extractive spans on QuoteSum and \textsc{Veri-Gran} (“L3.1” = LLaMa-3.1, “L3.2” = LLaMa-3.2). 
    For LLaMa-3.2-3B and Qwen3-8B, values are pre-softmax. “HS” = best single hidden-state layer baseline.}
    \label{tab:table_attribution_pre_post}
    \centering
      \begin{tabular}{@{}lcc@{}}
      \toprule
      Model & QuoteSum & \textsc{Veri-Gran} \\
      \midrule

      \arrayrulecolor{black!30}\specialrule{.5pt}{2pt}{2pt}
      L3.1-8B (all heads)        & 90.54 & 77.91 \\
      \rowcolor{mypink} L3.1-8B (post-softmax least RC, $k{=}20$) & 35.72 & 4.69 \\
      \rowcolor{mygreen} L3.1-8B (pre-softmax least RC, $k{=}20$) & 29.49 & 2.81 \\
      \rowcolor{mypink} L3.1-8B (post-softmax most RC, $k{=}20$) & 90.03 & 71.25 \\
      \rowcolor{mygreen} L3.1-8B (\textbf{pre-softmax most RC}, $k{=}20$) & \textbf{93.91} & \textbf{79.37} \\
      \specialrule{.5pt}{2pt}{2pt}\arrayrulecolor{black}

      \bottomrule
      \end{tabular}
\end{table}

\subsection{Qualitative Comparison of Attribution Strategies}
\label{sec:attribution_qualitative}

To better understand the effectiveness of various attention-based attribution strategies, we compare three different approaches using attention maps from Layer 15 of our model: (1) the mean attention across all heads, (2) the top-scoring head according to our attribution scoring technique, and (3) the worst-scoring head by the same measure. All methods were evaluated on the same input setup: a sales report document with the question ``\textit{What were the product sales on November 21st?}'' and the answer ``\textit{The product sales on November 21st were \$177.00.}''

Figure~\ref{fig:head-comparison} presents the attention heatmaps produced by each of the three strategies. The top-scoring head (Head 30, \cref{fig:tophead}) yields a sharply focused attribution map, precisely attending to tokens corresponding to the correct numerical value. In contrast, the mean attention across all heads (Figure~\ref{fig:meanhead}) produces a reasonable heatmap but also attends to several unrelated tokens, leading to less interpretable attributions. Finally, the worst-scoring head (Head 16, \cref{fig:bottomhead}) demonstrates diffuse and uninformative attention, highlighting mostly irrelevant tokens.

These observations qualitatively validate our scoring technique for identifying high-quality attribution heads and demonstrate that selectively using the best attention heads can significantly improve interpretability.

\section{Preliminaries}

Let $X = [\,x_{1}, x_{2}, \dots, x_{T}\,] \in \mathbb{R}^{T \times d}$ be the input sequence of length $T$, where each $x_{i} \in \mathbb{R}^{d}$ is an input token embedding and $d$ is the model's hidden dimension. For each attention head $h \in \{1, \dots, H\}$ in layer $\ell \in \{1, \dots, L\}$, define
\begin{equation}
\begin{aligned}
Q^{(\ell,h)} &= XW^{(\ell,h)}_{Q}, \quad
K^{(\ell,h)} = XW^{(\ell,h)}_{K}, \quad
V^{(\ell,h)} = XW^{(\ell,h)}_{V},
\end{aligned}
\end{equation}
where $Q^{(\ell,h)}, K^{(\ell,h)}, V^{(\ell,h)} \in \mathbb{R}^{T \times d_h}$ and $d_h = d / H$.

\paragraph{Attention Logits.}
The pre-softmax attention logits are given by
\begin{equation}
Z^{(\ell,h)} = Q^{(\ell,h)} (K^{(\ell,h)})^\top \in \mathbb{R}^{T \times T},
\quad
z^{(\ell,h)}_{i,j} = \langle q^{(\ell,h)}_i, k^{(\ell,h)}_j \rangle.
\end{equation}

\paragraph{Attention Weights.}
The post-softmax weights are
\begin{equation}
A^{(\ell,h)} = \mathrm{softmax}\left(\frac{Z^{(\ell,h)}}{\sqrt{d_h}}\right), \quad
a^{(\ell,h)}_{i,j} = \frac{\exp(z^{(\ell,h)}_{i,j} / \sqrt{d_h})}{\sum_{j'} \exp(z^{(\ell,h)}_{i,j'} / \sqrt{d_h})}.
\end{equation}

\paragraph{Attention Output.}
The output of the head is
\begin{equation}
Y^{(\ell,h)} = A^{(\ell,h)} V^{(\ell,h)} \in \mathbb{R}^{T \times d_h},
\end{equation}
and the full multi-head output is
\begin{equation}
\mathrm{MHA}^{(\ell)}(X) = \left[Y^{(\ell,1)} \,\Vert\, \dots \,\Vert\, Y^{(\ell,H)}\right] W_O^{(\ell)} \in \mathbb{R}^{T \times d}.
\end{equation}

We refer to $Z^{(\ell,h)}$ and $A^{(\ell,h)}$ as the pre-softmax affinities and post-softmax attention weights, respectively. While $A^{(\ell,h)}$ normalizes attention for token interaction, it also suppresses informative patterns in $Z^{(\ell,h)}$. Our work leverages $Z$ directly to extract statistical signals useful for contextual analysis, KV compression, and attribution.

\newpage

\begin{figure}
    \centering
    \includegraphics[width=\linewidth]{supp_figures/qmsum_contextualization_distribution.pdf}
    \caption{(see on screen) The distribution of RC upper bound (overlap area) for QmSum dataset. The first (last) row corresponds to heads in the first (last) layer. }
    \label{fig:rc_dist_qmsum}
\end{figure}

\begin{figure}
    \centering
    \includegraphics[width=\linewidth]{supp_figures/squad_contextualization_distribution.pdf}
    \caption{(see on screen) The distribution of RC upper bound (overlap area) for Squad v2 dataset. The first (last) row corresponds to heads in the first (last) layer. }
    \label{fig:rc_dis_squad}
\end{figure}
\begin{figure}
    \centering
    \includegraphics[width=\linewidth]{supp_figures/2WikiMultiHopQA_contextualization_distribution.pdf}
    \caption{(see on screen) The distribution of RC upper bound (overlap area) for 2WikiMultiHop dataset. The first (last) row corresponds to heads in the first (last) layer. }
    \label{fig:rc_dis_2wiki}
\end{figure}

\begin{figure}[t]
    \centering
    \begin{subfigure}{0.5\textwidth}
        \centering
        \includegraphics[page=1, trim=5cm 0cm 3cm 0cm, clip, width=\linewidth]{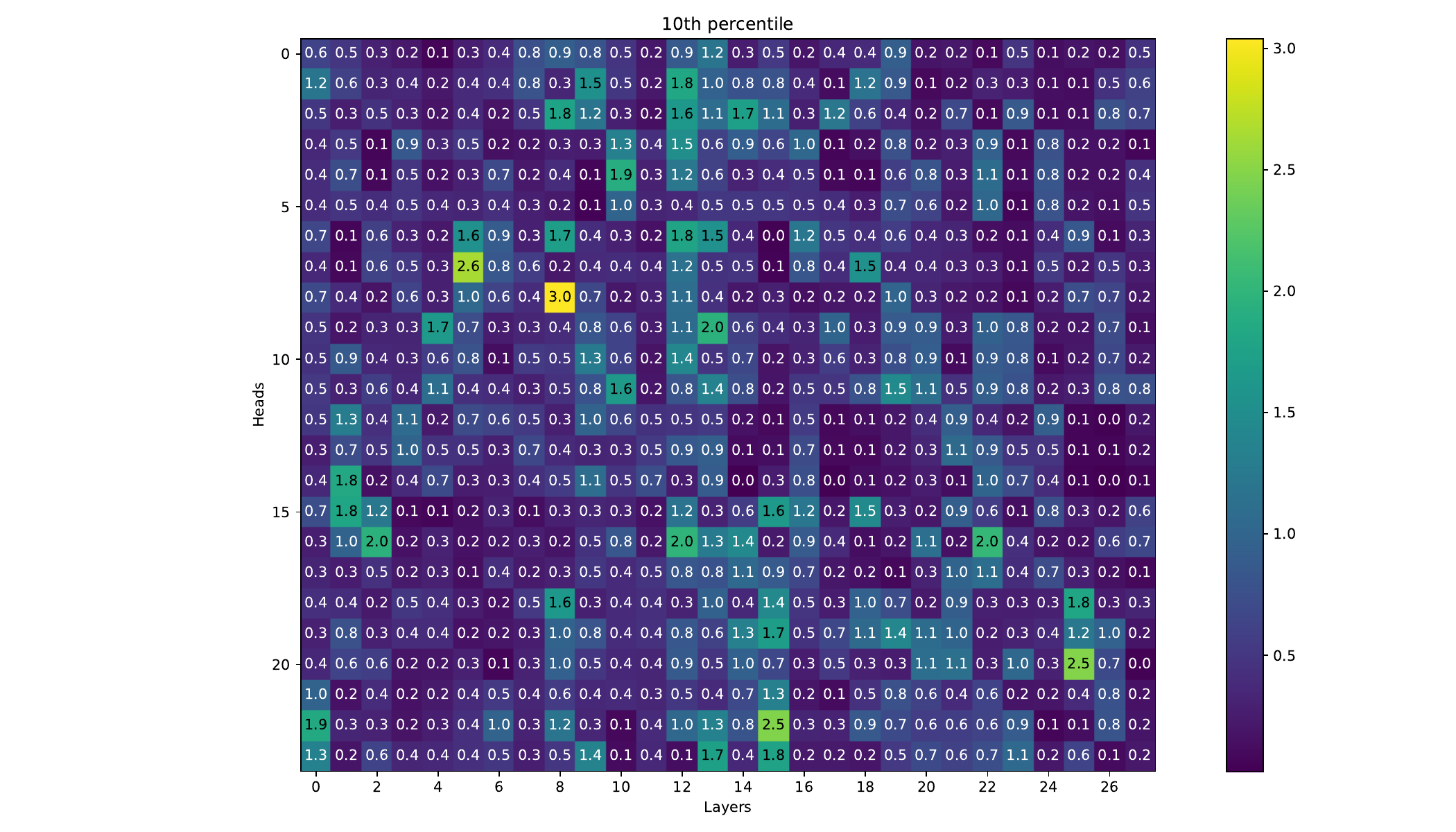}
    \end{subfigure}~
    \begin{subfigure}{0.5\textwidth}
        \centering
        \includegraphics[page=2, trim=5cm 0cm 3cm 0cm, clip, width=\linewidth]{supp_figures/qmsum_percentile_heatmaps.pdf}
    \end{subfigure}
    \begin{subfigure}{0.5\linewidth}
        \centering
        \includegraphics[page=3, trim=5cm 0cm 3cm 0cm, clip, width=\linewidth]{supp_figures/qmsum_percentile_heatmaps.pdf}
    \end{subfigure}~
    \begin{subfigure}{0.5\linewidth}
        \centering
        \includegraphics[page=4, trim=5cm 0cm 3cm 0cm, clip, width=\linewidth]{supp_figures/qmsum_percentile_heatmaps.pdf}
    \end{subfigure}
    \caption{\label{fig:percentile_rc_qmsum}(see on screen) Percentiles of RC upper bound (overlap area) across the QmSum dataset}
    \begin{subfigure}{0.5\textwidth}
        \centering
        \includegraphics[page=1, trim=5cm 0cm 3cm 0cm, clip, width=\linewidth]{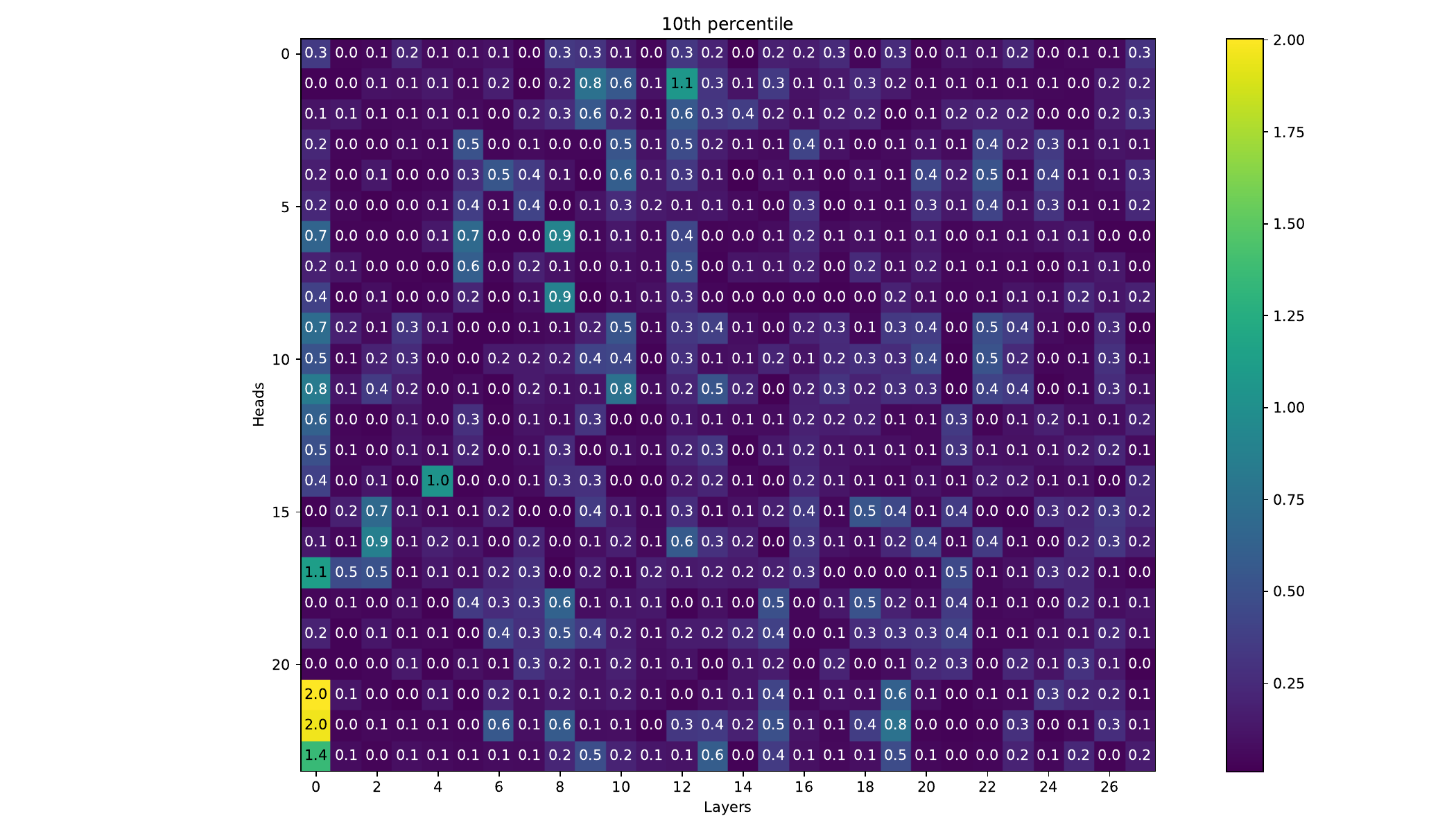}
    \end{subfigure}~
    \begin{subfigure}{0.5\textwidth}
        \centering
        \includegraphics[page=2, trim=5cm 0cm 3cm 0cm, clip, width=\linewidth]{supp_figures/squad_percentile_heatmaps.pdf}
    \end{subfigure}
    \begin{subfigure}{0.5\linewidth}
        \centering
        \includegraphics[page=3, trim=5cm 0cm 3cm 0cm, clip, width=\linewidth]{supp_figures/squad_percentile_heatmaps.pdf}
    \end{subfigure}~
    \begin{subfigure}{0.5\linewidth}
        \centering
        \includegraphics[page=4, trim=5cm 0cm 3cm 0cm, clip, width=\linewidth]{supp_figures/squad_percentile_heatmaps.pdf}
    \end{subfigure}
    \caption{\label{fig:percentile_rc_squad}(see on screen) Percentiles of RC upper bound (overlap area) across the Squad dataset}
\end{figure}

\begin{figure}[t]
    \centering
    \begin{subfigure}{0.5\textwidth}
        \centering
        \includegraphics[page=1, trim=5cm 0cm 3cm 0cm, clip, width=\linewidth]{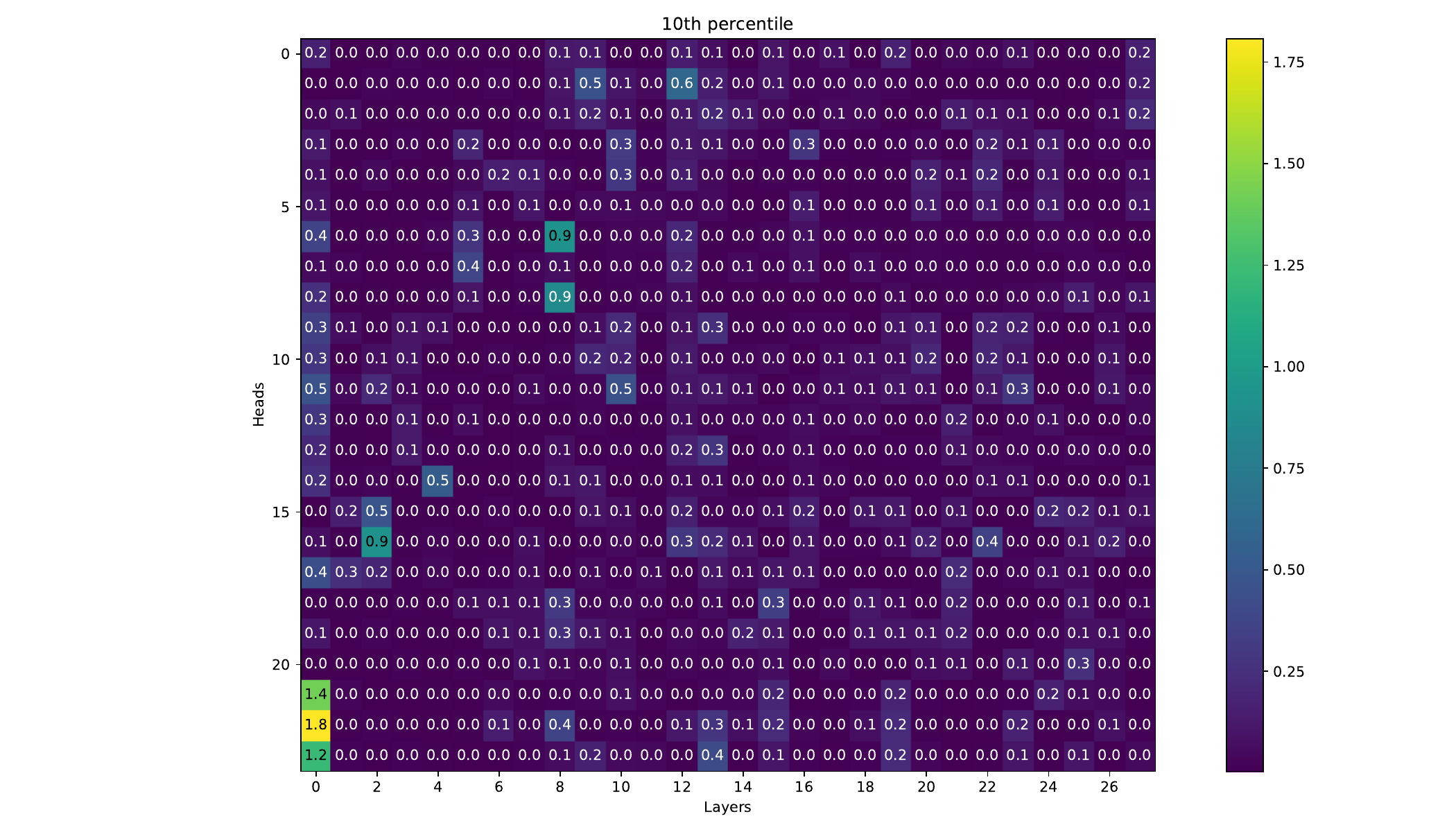}
    \end{subfigure}~
    \begin{subfigure}{0.5\textwidth}
        \centering
        \includegraphics[page=2, trim=5cm 0cm 3cm 0cm, clip, width=\linewidth]{supp_figures/2WikiMultiHopQA_percentile_heatmaps.pdf}
    \end{subfigure}
    \begin{subfigure}{0.5\linewidth}
        \centering
        \includegraphics[page=3, trim=5cm 0cm 3cm 0cm, clip, width=\linewidth]{supp_figures/2WikiMultiHopQA_percentile_heatmaps.pdf}
    \end{subfigure}~
    \begin{subfigure}{0.5\linewidth}
        \centering
        \includegraphics[page=4, trim=5cm 0cm 3cm 0cm, clip, width=\linewidth]{supp_figures/2WikiMultiHopQA_percentile_heatmaps.pdf}
    \end{subfigure}
    \caption{\label{fig:percentile_rc_2wiki}(see on screen) Percentiles of RC upper bound (overlap area) across the 2WikiMultiHop dataset}
\end{figure}

\begin{figure}[t]
    \centering
    \begin{subfigure}[b]{0.45\textwidth}
        \includegraphics[width=\textwidth]{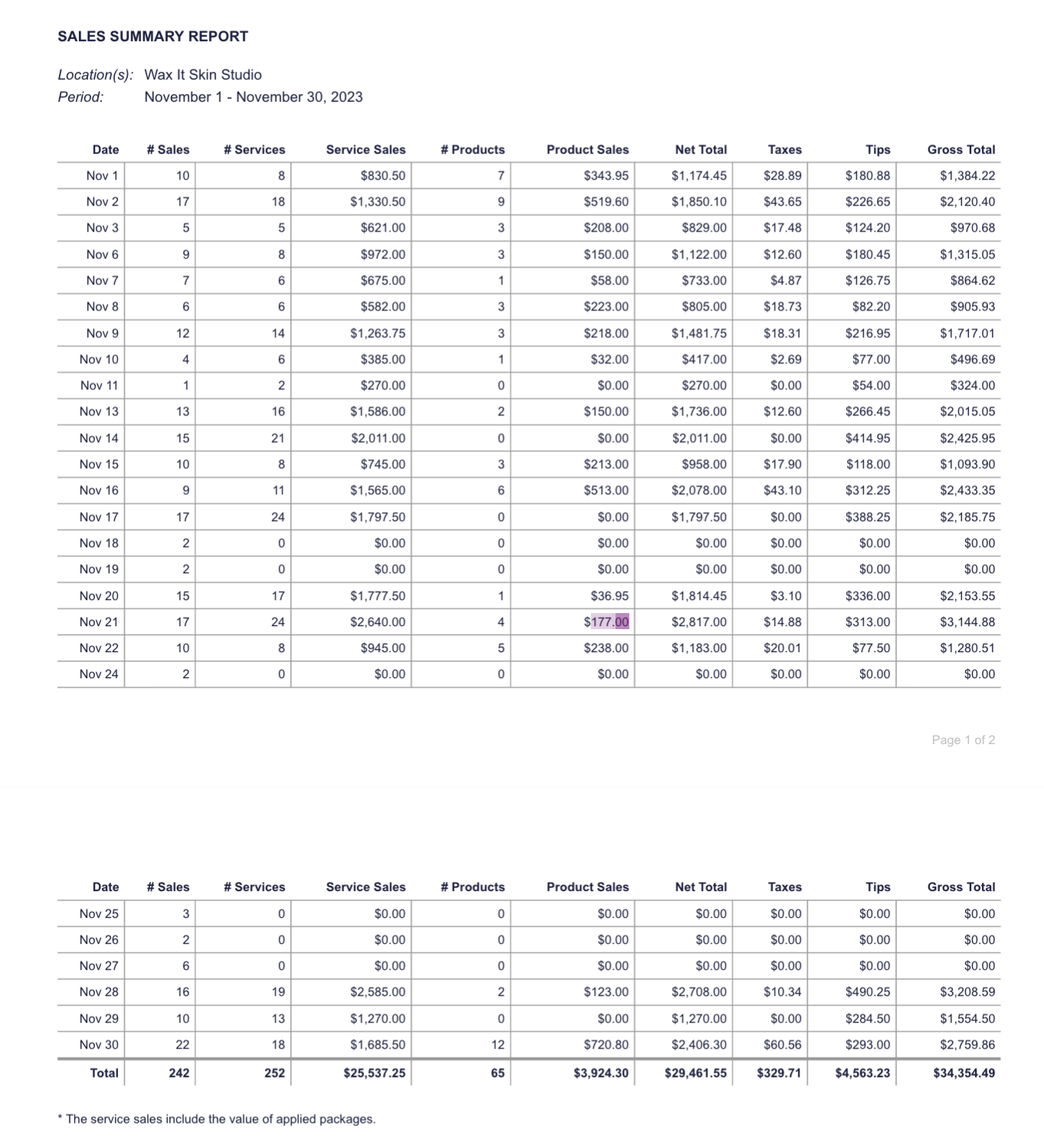}
        \caption{Top-scoring head (Head 30)}
        \label{fig:tophead}
    \end{subfigure}
    \hfill
    \begin{subfigure}[b]{0.45\textwidth}
        \includegraphics[width=\textwidth]{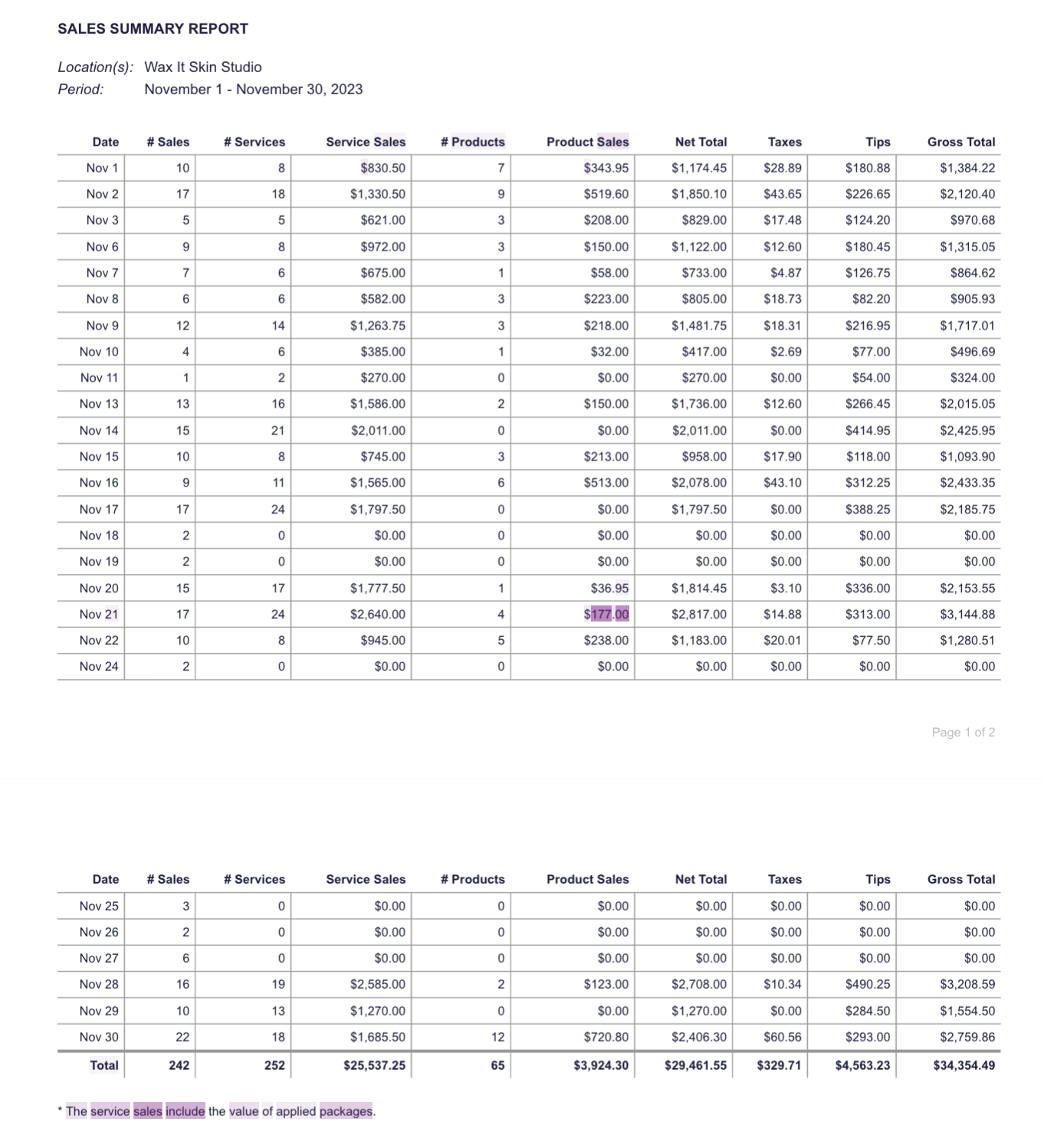}
        \caption{Mean of all heads}
        \label{fig:meanhead}
    \end{subfigure}
    
    \vspace{1em}
    
    \begin{subfigure}[b]{0.6\textwidth}
        \includegraphics[width=\textwidth]{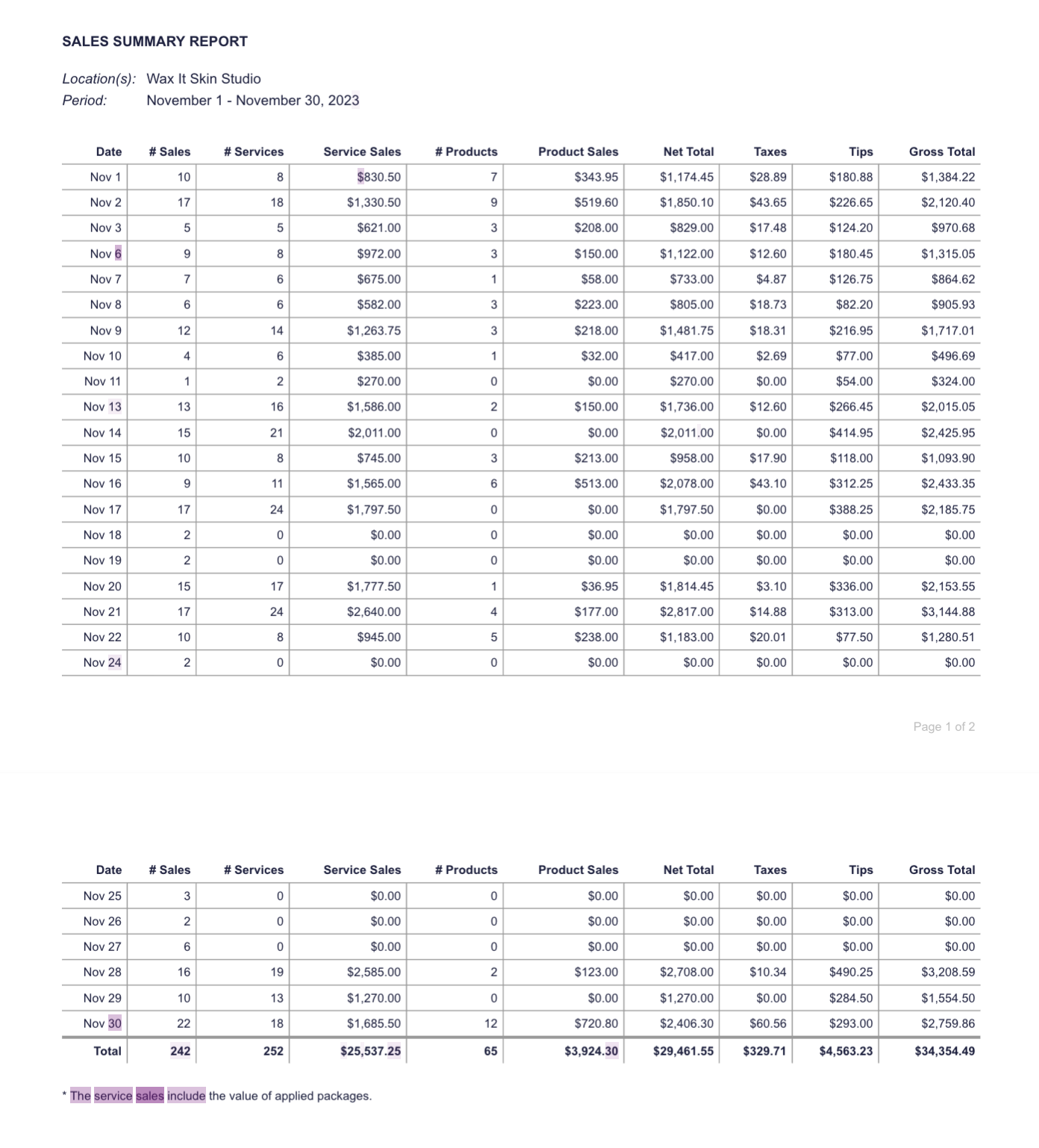}
        \caption{Worst-scoring head (Head 16)}
        \label{fig:bottomhead}
    \end{subfigure}
    
    \caption{Attention heatmaps from Layer 15 using three attribution strategies. The top head yields focused and accurate attribution, the mean head shows diluted but somewhat relevant attention, and the worst head highlights largely irrelevant regions.}
    \label{fig:head-comparison}
\end{figure}

\end{document}